\documentclass[conference]{IEEEtran}
\usepackage{times}

\usepackage[numbers]{natbib}
\usepackage{multicol}
\usepackage[bookmarks=true]{hyperref}
\usepackage{graphics} 
\usepackage{epsfig} 
\usepackage{amsmath}
\usepackage{mathptmx} 
\DeclareMathAlphabet{\mathcal}{OMS}{cmsy}{m}{n}
\DeclareSymbolFont{largesymbols}{OMX}{cmex}{m}{n}
\usepackage{amssymb}  
\usepackage[T1]{fontenc}
\usepackage{subfigure}
\usepackage{bm}
\usepackage{hyperref}
\usepackage{booktabs}
\usepackage{makecell}
\usepackage{tabularx}
\usepackage{multirow}
\usepackage{threeparttable}
\usepackage{color, colortbl}
\definecolor{Gray}{gray}{0.9}
\usepackage{capt-of}

\IEEEoverridecommandlockouts 
\begin{document}

\title{
Quadric Representations for \\ LiDAR Odometry, Mapping and Localization
}

\author{Chao~Xia*,
        Chenfeng~Xu*,
        Patrick~Rim,
		Mingyu Ding,
        Nanning Zheng$^\dag$
       \\ Kurt Keutzer,
        Masayoshi Tomizuka, and
        Wei Zhan
       
\thanks{C. Xia and N. Zheng are with the Institute of Artificial Intelligence and Robotics, Xi’an Jiaotong University, Xi’an, Shaanxi 710049, P.R. China; Email: {\tt\small xc06210417@stu.xjtu.edu.cn; nnzheng@mail.xjtu.edu.cn}}
\thanks{C. Xu, M. Ding, K. Keutzer, M. Tomizuka and W. Zhan are with the University of California, Berkeley, USA; Email: {\tt\small \{xuchenfeng, myding, keutzer, tomizuka, wzhan\}@berkeley.edu}}
\thanks{P. Rim is with the California Institute of Technology, USA; Email: {\tt\small prim@caltech.edu}}
\thanks{$^\dag$N. Zheng is the corresponding author.}
\thanks{$^*$C. Xia and C. Xu contributed equally to this work.}
}

\maketitle

\begin{abstract}
Current LiDAR odometry, mapping and localization methods leverage point-wise representations of 3D scenes and achieve high accuracy in autonomous driving tasks. However, the space-inefficiency of methods that use point-wise representations limits their development and usage in practical applications. In particular, scan-submap matching and global map representation methods are restricted by the inefficiency of nearest neighbor searching (NNS) for large-volume point clouds. To improve space-time efficiency, we propose a novel method of describing scenes using quadric surfaces, which are far more compact representations of 3D objects than conventional point clouds. In contrast to point cloud-based methods, our quadric representation-based method decomposes a 3D scene into a collection of sparse quadric patches, which improves storage efficiency and avoids the slow point-wise NNS process. Our method first segments a given point cloud into patches and fits each of them to a quadric implicit function. Each function is then coupled with other geometric descriptors of the patch, such as its center position and covariance matrix. Collectively, these patch representations fully describe a 3D scene, which can be used in place of the original point cloud and employed in LiDAR odometry, mapping and localization algorithms. We further design a novel incremental growing method for quadric representations, which eliminates the need to repeatedly re-fit quadric surfaces from the original point cloud. Extensive odometry, mapping and localization experiments on large-volume point clouds in the KITTI and UrbanLoco datasets demonstrate that our method maintains low latency and memory utility while achieving competitive, and even superior, accuracy.
\end{abstract}

\IEEEpeerreviewmaketitle

\section{INTRODUCTION}

Real-time odometry, mapping and localization in perceptually-challenging environments are critical capabilities for robotics and autonomous driving vehicles. Owing to the high accuracy and long-range detection ability of 3D Light Detection and Ranging (LiDAR), recent odometry~\cite{li2019net,shan2018lego,wang2021f,zheng2021efficient,zhang2015visual}, mapping,~\cite{zhang2014loam,deschaud2018imls,liu2021balm,koide2021globally,cvivsic2022soft2} Simultaneous Localization and Mapping (SLAM)~\cite{hess2016real,dellenbach2022ct,neuhaus2019mc2slam}, and localization~\cite{levinson2010robust,levinson2007map,xia2022onboard,chen2021range} methods have made notable progress. Compared to vision-based methods~\cite{mur2015orb,mur2017orb,campos2021orb,qin2018vins}, LiDAR-based methods achieve higher accuracy and better robustness to adverse illumination effects. Despite the promising performance of recent point-wise LiDAR-based methods, they suffer from 1) the restrictive memory constraints due to the high volume required to store the global map and 2) the heavy overhead of point-wise nearest neighboring searching (NNS), especially on tasks that require large-volume point-clouds, such as mapping and localization. 

\begin{figure}[t]
\centering
    \subfigure[Original point cloud]{\includegraphics[height=3.3cm]{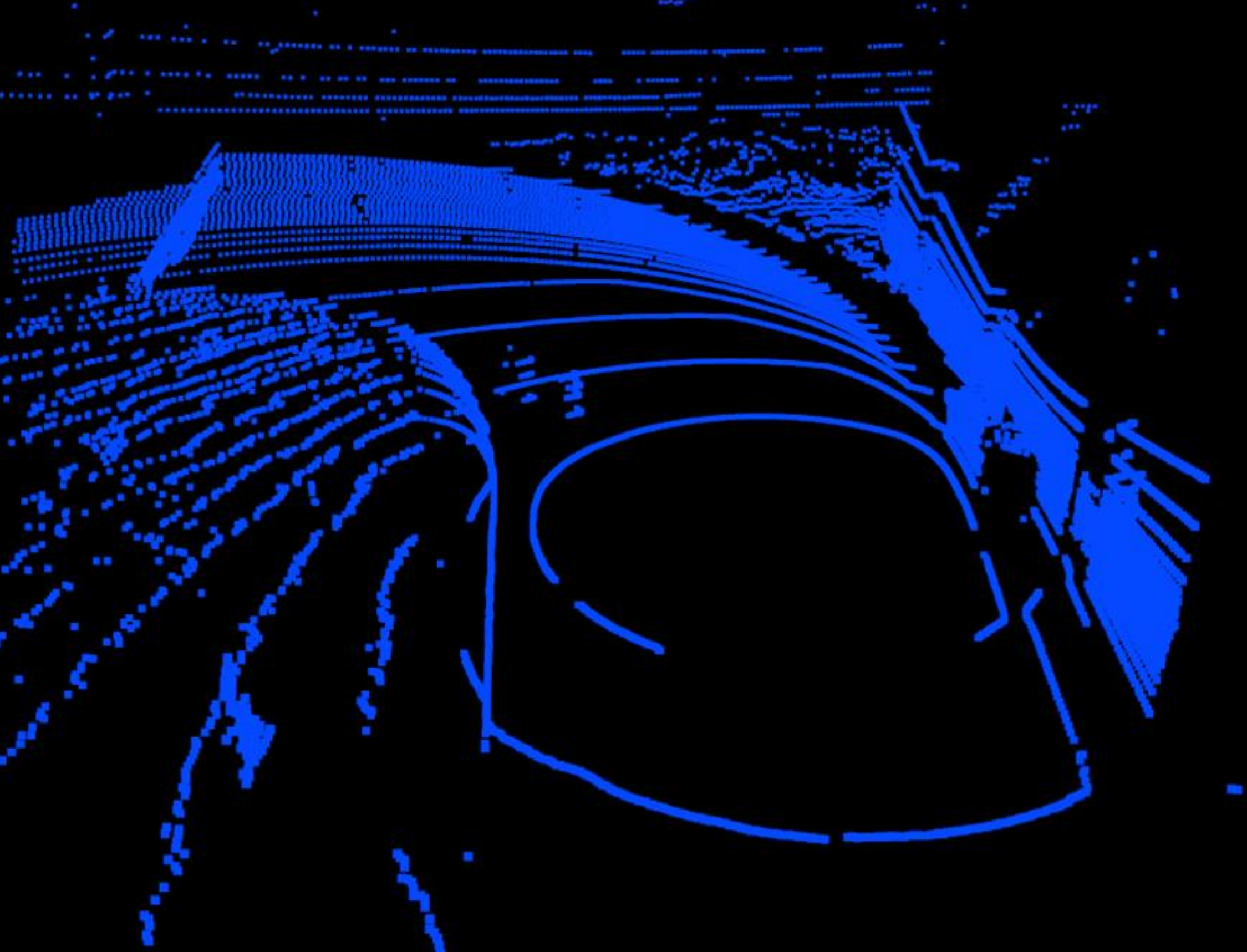}}
    \subfigure[Normal distribution representation]{\includegraphics[height=3.3cm]{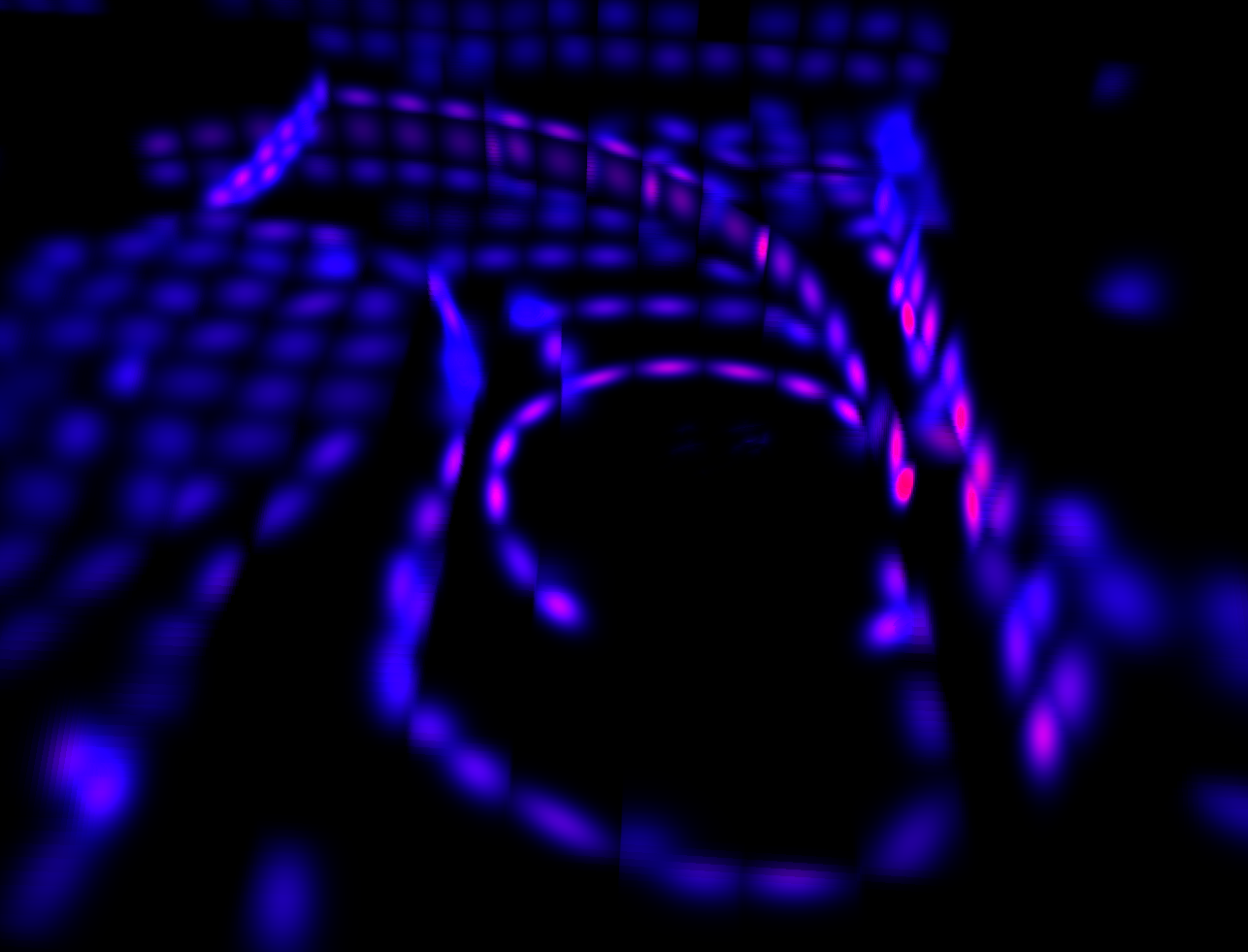}}\\
    \subfigure[Manhattan representation]{\includegraphics[height=3.3cm]{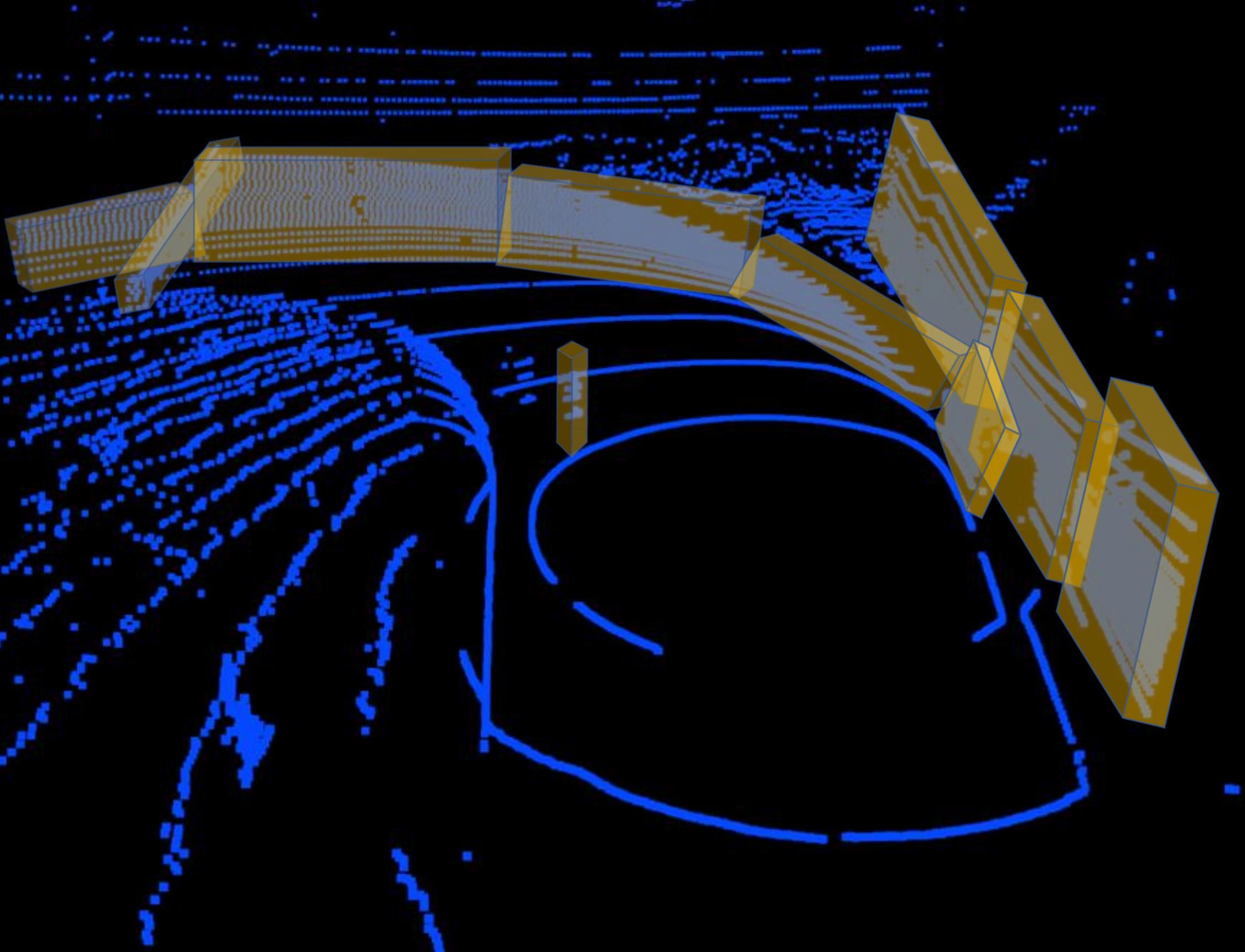}}
    \subfigure[Quadric representation]{\includegraphics[height=3.3cm]{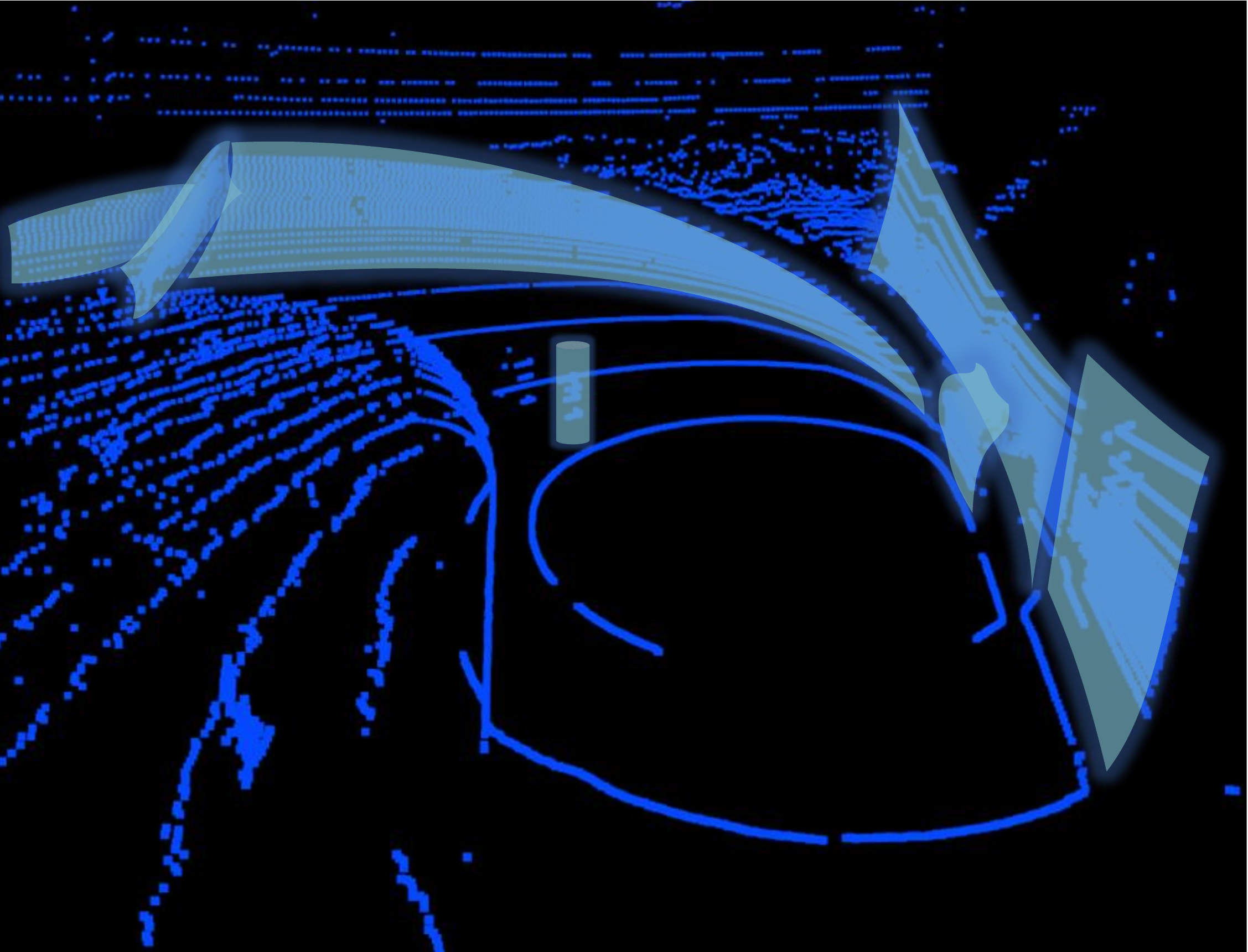}}
    \caption{Different representations of a 3D scene. The depicted example is an entrance of an underground parking lot.}
    \vspace{-8pt}
    \label{representations}
\end{figure}

To tackle these challenges, more compact scene (map) representation methods were proposed~\cite{levinson2007map}. For example, voxelization is the most common point cloud down-sampling method (see Fig.~\ref{representations} (a)), and it is usually combined with the normal distribution method~\cite{ndt,xia2022onboard} that represents a scene using normal distribution parameters for each voxel, as shown in Fig.~\ref{representations} (b). However, these voxel and normal distribution representations are prone to losing fine-grained information in the 3D scenes, and thus they result in unsatisfactory performance when used. Previous works also introduce the Manhattan representation of scenes~\cite{behley2018efficient, zhang2014loam, segal2009generalized}. An example of the Manhattan representation is shown in Fig. \ref{representations} (c). However, real 3D scenes are full of curved and rounded objects, meaning that the Manhattan representation cannot sufficiently describe 3D scenes, such as curving roads and walls.

\begin{figure*}[t]
	\centering
	\includegraphics[width=.95\textwidth]{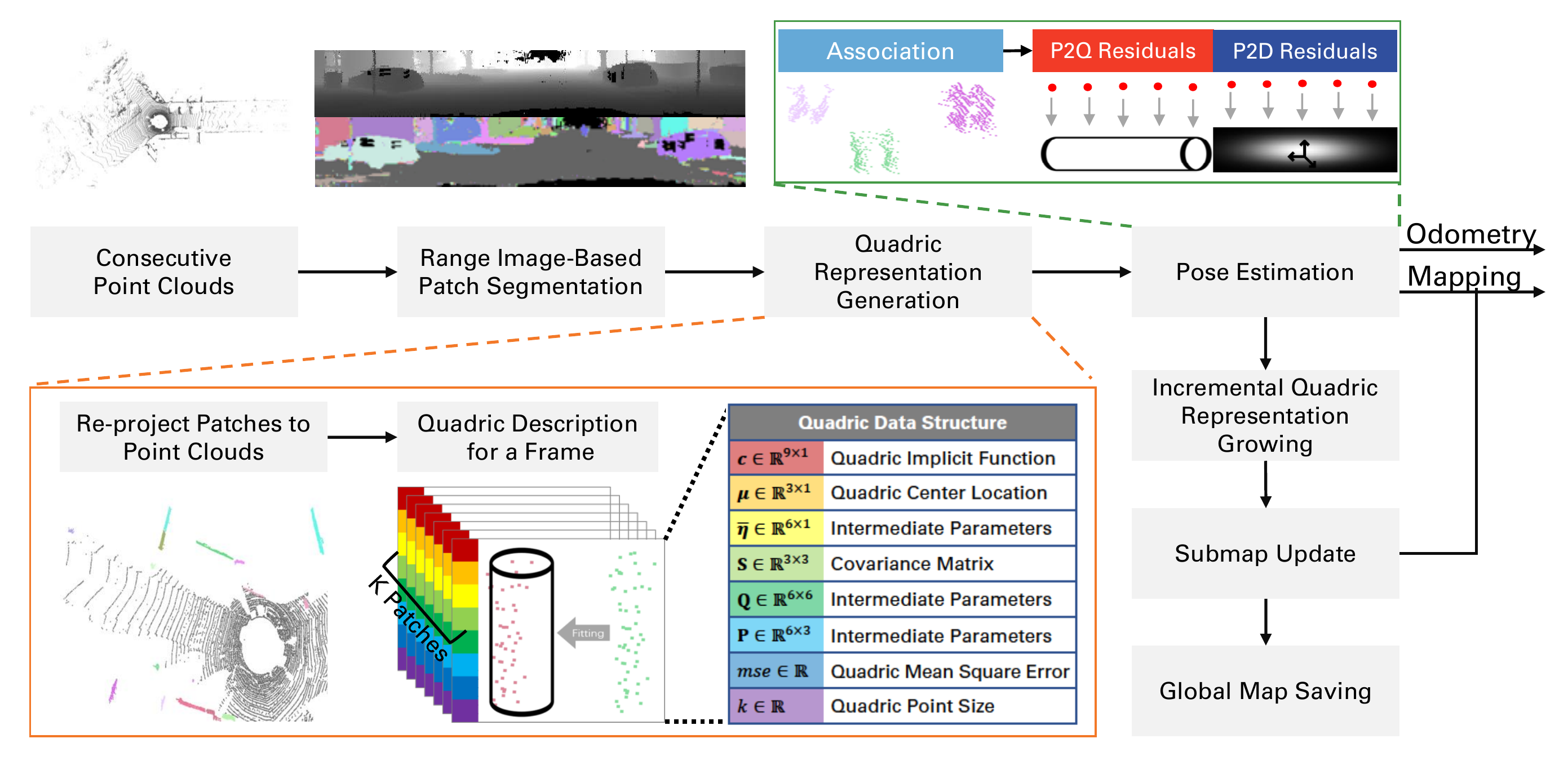}
        \vspace{-8pt}
	\caption{\textbf{Quadric Representation Overview.} Our framework can be applied to LiDAR odometry, mapping and localization tasks, where the LiDAR pose will be estimated by optimizing the point cloud-to-quadric representation (P2Q) or point cloud-to-degenerated quadric representation (P2D) projection distance. Using our incremental growing method, the quadric surfaces will grow as new points are sampled. Our space-efficient quadric representation can be utilized for scan-submap matching (mapping) and scan-map matching (localization) in place of dense point-wise representations.}
        \vspace{-4pt}
	\label{framework}
\end{figure*}

As an alternative, we propose an efficient quadric representation~\cite{vaskevicius2010extraction} of scenes that can be used for odometry, mapping, and localization tasks. Compared to dense point-wise representations, our quadric representation is a sparse representation that decomposes 3D scenes into a collection of quadric surfaces, each of which is represented by a quadric implicit function. As opposed to voxel and Manhattan representations, the quadric representations are more compact and fit more closely to real 3D scenes, as shown in Fig. \ref{representations} (d). The overview of our proposed quadric representation-based LiDAR odometry approach is depicted in Fig.~\ref{framework}. Firstly, the raw 3D point cloud from a LiDAR scan is projected onto a spherical range image to facilitate patch segmentation~\cite{bogoslavskyi2016fast}. Then, quadric implicit functions are fitted on the extracted patches in an optimal manner, where multi-thread parallel computing is applied for acceleration. For each patch, the center location, covariance matrix, and other geometric information are calculated and saved with the patch's quadric implicit function. Finally, the quadric patches replace the original point cloud to perform scan-scan matching (odometry). In addition, we also propose an incremental quadric growing method for submap and global map generation, which allows quadric surfaces to be re-used in mapping (scan-submap matching) and localization (scan-global map matching) tasks, avoiding the slow process of repeated generation.

We perform extensive experiments on the KITTI and UrbanLoco datasets to demonstrate that even as the volume of point clouds increases, our proposed quadric representation-based framework is able to maintain low latency and memory utility, while still achieving competitive or superior performance, for odometry, mapping and localization tasks. Specifically, our proposed method achieves better performance in odometry task (scan-scan) in terms of Average Transition Error (ATE) and Average Orientation Error (ARE) with similar latency relative to the current state-of-the-art method \cite{zhang2014loam}. For mapping tasks, our proposed quadric representation method can run at 10 HZ with comparable accuracy since there is no point-wise NNS and less residuals for pose estimation. For localization tasks, the quadric representation can greatly reduces the storage occupancy for global map storing compared with point cloud from 16.9 GB to 11.2 MB for a real-world global map, where each quadric center location is encoded to a binary code for fast local map searching.
To summarize, the main contributions of this paper are as follows:
\begin{itemize}
    \item We propose a quadric representation-based method that approximately represents a scene, where surfaces from point-cloud representations of 3D scenes are extracted and fitted to a different type of quadrics. These quadric-based representations can be used in place of the original point clouds for LiDAR odometry, mapping and localization. Our representation significantly reduces the space needed for global map storage and improves the time-efficiency of feature association, and maintains current accuracy benchmarks when used in scan-scan and scan-map registration tasks.
    
    \item We propose an incremental growing method for quadric representations, which speeds up submap and map generation. This method allows for the gradual expansion of quadric surfaces, eliminating the need for the computationally expensive task of repeatedly re-fitting quadratic surface implicit functions from the original point cloud.

    \item Through experiments on the KITTI~\cite{Geiger2012CVPR} and UrbanLoco~\cite{wen2020urbanloco} datasets, we demonstrate that our method provides a significant improvement in efficiency, as well as promising performance on map compression, LiDAR odometry, mapping, and localization tasks.
\end{itemize}

\section{RELATED WORK}
LiDAR-based odometry, mapping and localization are challenging tasks that require the integration of various techniques from computer vision, robotics, and machine learning. Different kinds of scene representation methods demonstrate different advantages.

Point-wise representation methods achieve high accuracy for odometry and mapping tasks. For example, LOAM~\cite{zhang2014loam}, uses a feature-based registration between consecutive scans, and a least-squares optimization to compute the motion of the LiDAR. It is currently the best-performing LiDAR SLAM (Simultaneous Localization and Mapping) algorithm on the KITTI odometry benchmark. Similarly, MULLS~\cite{pan2021mulls} extracts roughly classified feature points (ground, facade, pillar, beam, etc.) from each frame, and then uses joint linear least square optimization in these geometric feature points. LeGO-LOAM~\cite{shan2018lego} applies an image-based segmentation method~\cite{bogoslavskyi2016fast} to the range image to group points into many clusters. Then, it finds correspondences for the same object and narrows down potential candidates for correspondences, improving matching accuracy. It also uses a pose graph to create the surrounding submap for scan-submap matching. LIO-SAM~\cite{shan2020lio}, a tightly coupled optimization framework fusing LiDAR odometry and GPS/IMU measurements, improves SLAM accuracy and robustness with a multi-sensor fusion system. 

Besides the geometry-based method discussed above, 3D feature registration~\cite{guo2013rotational,rusu2009fpfh} and deep-learning-based methods~\cite{ao2021spinnet, zheng2020lodonet, li2019net} are built on the principle of extracting discriminative descriptors from a given point cloud and then estimating a rigid transformation matrix using the Iterative Closest Point (ICP) algorithm. These methods can achieve high compression ratios but have heavy overhead, such that they cannot be applied to LiDAR localization tasks in large-scale scenarios due to the large volume of memory required for global map storage and the time-inefficiency of finding feature correspondences. Even using an efficient KD-tree, the time complexity of the NNS algorithm used for correspondence searching is still $O(log(n))$. 

Due to these issues of time and space inefficiency, many novel scene representation methods were proposed to reduce the space required to store the map while ensuring that using a compressed representation could maintain sufficient accuracy. Mesh-based reconstruction methods \cite{ZhangRSS22} are proposed to increase point cloud density while reducing its storage and transmitting size. Gaussian distributions as volumetric primitives \cite{dhawale2020efficient} present scenes accurately and efficiently for dense multi-fidelity 3D mapping. Surfel-based methods~\cite{behley2018efficient} estimate a robot’s pose by exploiting the projective data association between the current scan and a rendered model view from a surfel map. Poisson surface-based methods~\cite{vizzo2021poisson,kazhdan2013screened} represent maps as a triangle mesh computed via Poisson surface reconstruction and achieve accurate registration of the scans with the model using a novel frame-to-mesh registration algorithm. But triangular-mesh-based representation methods are usually inefficient in single frame processing due to the complex model fitting. Hierarchical map-building and encoding algorithms~\cite{xia2022onboard} divide a given LiDAR scan into voxels and calculate the normal distribution parameters for the valid voxels, and an improved Normal Distribution Transform algorithm is used for pose estimation. 

In this paper, we propose the novel quadric representation that is efficient and fits well to various 3D scenes. Our novel method leverages quadric representations to improve space-time efficiency for LiDAR odometry, mapping and localization tasks without trading off accuracy. 

\section{NOTATION AND PRELIMINARIES}
\subsection{Notation}
We will estimate the transformation matrix,
$${\bf T}=\begin{bmatrix} {\bf R} & \boldsymbol{t} \\ {\boldsymbol 0}^{\mathrm{T}} & 1 \end{bmatrix} \in \mathbb{SE}(3)|{\bf R} \in \mathbb{SO}(3),\boldsymbol{t}\in \mathbb{R}^{3\times 1}\text{,}$$
which aligns the source point cloud $\mathcal{M}=\{\boldsymbol{m}_j = [x, y, z]|j=1,2,\cdots,N_m\}$ and target point cloud ${\mathcal N} = \{\boldsymbol{n}_j = [x, y, z]|j=1,2,\cdots,N_{n}\}$. We use $\boldsymbol{n} = {\bf T} \boldsymbol m = {\bf R} \boldsymbol m + \boldsymbol t$ to indicate a coordinate transformation from a point $\boldsymbol m$ to a point $\boldsymbol{n}$ using a transformation matrix $\bf T$. We define $\operatorname f(\boldsymbol c, \boldsymbol q) = \boldsymbol {c}^T \boldsymbol q = 0$ to be the quadric implicit function, where $\boldsymbol c = [c_0, c_1, \cdots, c_9]^T$ and $\boldsymbol q = [x^2, y^2, z^2, xy, yz, xz, x, y, z, 1]^T$. $\boldsymbol q$ can be separated into three parts: the quadratic term $\boldsymbol \eta = [x^2, y^2, z^2, xy, yz, xz]^T$, the linear term $\boldsymbol p = [x, y, z]^T$ and the constant term. The quadratic coefficient vector can be separated into $\boldsymbol c_\eta=[c_0,\cdots,c_5]^T$, $\boldsymbol c_p
=[c_6,c_7,c_8]^T$, and $c_9$ correspondingly. Then, we can rewrite the quadric implicit function as $\operatorname{f}(\boldsymbol c, \boldsymbol q)=\boldsymbol c_\eta^T \boldsymbol \eta+\boldsymbol c_p^T \boldsymbol p +c_9$. In matrix form, we can write $\operatorname f(\boldsymbol c, \boldsymbol q) = \boldsymbol {p}^T \bm A \boldsymbol p + \bm {b}^T \boldsymbol p + c = 0$, where:

$$\bm A = \begin{bmatrix} c_0 & c_3/2 & c_5/2 \\ c_3/2 & c_1 & c_4/2 \\ c_5/2 & c_4/2 & c_2 \end{bmatrix}\text{, }\bm b = [c_6, c_7, c_8]^T\text{, }c = c_9\text{.}$$

As defined by Taubin~\cite{taubin1991estimation}, the distance from a point to a quadric surface is:

\begin{equation}
    \operatorname {d_q}(\boldsymbol c, \boldsymbol p) \approx \operatorname s(\boldsymbol c, \boldsymbol q) = \frac{\operatorname f^2(\boldsymbol c, \boldsymbol q)}{\operatorname g^2(\boldsymbol c, \boldsymbol q)}
\end{equation}

where $\operatorname g^2(\boldsymbol c, \boldsymbol q) = \left| \nabla_{\boldsymbol p}\operatorname f(\boldsymbol c, \boldsymbol q))\right|^2$.

\subsection{Quadric Representation}

We propose a quadric representation of 3D scenes. As shown in Fig. \ref{framework}, we first project the point cloud as a spherical range image, and then we utilize the range-map segmentation algorithm \cite{bogoslavskyi2016fast} to classify the label for each point. Afterward, we apply the quadric fitting to the patches and we use a threshold to select the fitted quadric patches based on mean square error (MSE). If the corresponding MSE does not exceed the threshold, we preserve the valid patches with its MSE ($mse$), point size ($k$), implicit function ($\boldsymbol c$), center location ($\boldsymbol \mu$), covariance matrix ($\mathbf{S}$), and other intermediate information matrices such as the quadratic center $\overline{\boldsymbol \eta}$, $\mathbf{Q}$, and $\mathbf{P}$, as defined in~\cite{vaskevicius2010extraction}. The data structure of the quadric representation is shown in the bottom part of Fig.~\ref{framework}. Experiments demonstrate that tens (up to hundreds) of quadric surfaces can be extracted from a frame of a LiDAR scan in common urban scenes and that all of them are important basic primitives for scan-scan and scan-submap matching.

However, due to measurement error, segmented surfaces with an MSE that exceeds the threshold cannot be successfully fitted to a quadric. In addition, regular flat planes and curved surfaces are not as common in unstructured scenes such as countryside roads as they are in man-made scenes. To improve the registration robustness and avoid mismatching, ``invalid patches'' (patches that fail to be fitted to a quadric) will also be saved with the same information, which we call degenerated quadric surface. In a degenerated quadric surface, a zero vector is assigned as the implicit function, and we propose to approximate the surface with a normal distribution, in which we keep the center location and covariance matrix. We then apply them to the point-distribution residual generation and registration procedure manually.

\section{METHODOLOGY}
It is quite common to use point-line-plane features for LiDAR odometry, mapping and localization. However, point-wise operations are time-consuming. Intuitively, the environment description obtained by LiDAR scanning is actually a surface sampling process. If all the surfaces can be extracted and fitted to accurate implicit functions, then the scanning points of the next frame are projected to the corresponding surfaces such that inter-frame pose estimation can be conducted by minimizing the projection distance.

While in theory, an infinite order polynomial could describe a surface with arbitrary accuracy, in practice, this is computationally infeasible and may lead to poor generalization and poor performance on new, unseen data. Thus, it is more common to use lower-order polynomials or other types of surface representations that provide a good trade-off between accuracy and computational feasibility. We choose to use quadrics to describe and represent surfaces, which can be considered an extension of linear features such as points, lines, and planes.

Quadrics can approximate most surface features, even when they are incompletely sampled, or only a small part of them is visible. For example, walls, the ground, tree trunks, tree crowns, and other structures that frequently appear in traffic scenes can be approximated using different types of quadrics. We show that quadrics preserve enough information for accurate pose estimation. Moreover, the quadric fitting process is quick and efficient, and its time complexity only grows incrementally. This is extremely beneficial, especially in the process of map/submap generation.

In the following section, we demonstrate how to apply the quadric representation-based pose estimation to LiDAR odometry (scan-scan matching), mapping (scan-submap matching), and localization (scan-map matching). We will derive the pose Jacobian matrix from the point-to-patch distance and present the quadric association and quadric growing methods.

\subsection{Quadric Representation-Based Point Cloud Registration}
Quadric association is the basis of scan-scan registration. Assume that we have already obtained the quadric correspondences, where each quadric surface from the source scan is perfectly corresponding to a quadric surface from the target scan. 
Optimizing a point transformation matrix between two consecutive LiDAR scans, the residual of the projected distance between a point and its corresponding quadric is
\begin{equation}
    res_i^q = \operatorname{d_q}(\boldsymbol c_j, \bf {T} \boldsymbol p_i)
    \label{resq}\text{.}
\end{equation}

The residual of the projected distance between a point and its corresponding degenerated quadric surface is
\begin{equation}
    res_i^d = \operatorname{d_d}(\mathcal{N}_j, \mathbf{T} \boldsymbol {p}_i) = \left(\mathbf{T}\boldsymbol {p}_i- \boldsymbol{\mu}_j \right)^T \mathbf{S}_j^{-1} \left(\mathbf{T}\boldsymbol {p}_i-\boldsymbol{\mu}_j \right)
    \label{resd}\text{.}
\end{equation}

Stacking all of the residuals $\boldsymbol {res} = [\boldsymbol {res}^q;\boldsymbol {res}^d]$, where $\boldsymbol {res}^q = [res^q_1, res^q_2, \cdots , res^q_m]^T$ and $\boldsymbol {res}^d = [res^d_1, res^d_2, \cdots, res^d_n]^T$, a nonlinear cost function can be obtained. A Lie algebra perturbation model is used to compute the Jacobian matrix of Eq.~\ref{resq} with respect to $\bf T$, as follows:
\begin{equation}
    \mathbf {J}^q = \frac{\partial res_i^q}{\partial \boldsymbol \xi}=\frac{\partial res_i^q}{\partial (\mathbf{T} \boldsymbol{p}_i)}\frac{\partial (\mathbf{T} \boldsymbol{p}_i)}{\partial \boldsymbol \xi} =  \frac{2 \operatorname{f} \cdot \operatorname{f}^\prime \cdot \operatorname{g}^2-2\operatorname{g}^T \cdot \operatorname{g}^\prime \cdot \operatorname{f}^2}{\operatorname{g}^4}\text{,}
\end{equation}
where $\boldsymbol \xi \in \mathbb{R}^{6 \times 1}$ is the Lie algebra vector with respect to the transformation matrix, $\operatorname{f}^\prime = \operatorname{f}^\prime\left(\boldsymbol{c}_j,\bf T \boldsymbol{p}_i\right)=[2 {\bf A} \cdot {\bf T} \boldsymbol p_i + \boldsymbol b]^T \cdot {\bf T}_p^\prime$, and $\operatorname{g}^\prime = \operatorname{g}^\prime\left( \boldsymbol{c}, {\bf T} \boldsymbol{p}_i\right)=2 {\bf A} \cdot {\bf T}_p^\prime$. ${\bf T}_p^\prime$ is the perturbation increment in Lie algebra:
\begin{equation}
    {\bf T}_p^\prime = \frac{\partial (\bf{T} \boldsymbol p_i)}{\partial \boldsymbol \xi}=\left[\bf {I}, -(\bf{R} \boldsymbol p_i + \boldsymbol{t})^\wedge \right]\text{.}
\end{equation} 

The Jacobian matrix of Eq.~\ref{resd} with respect to $\bf T$ is
\begin{equation}
    \mathbf {J}^d = 2\left(\mathbf{T}\boldsymbol {p}- \boldsymbol{\mu} \right)^T \mathbf{S}^{-1} \cdot {\bf T}_p^\prime\text{.}
\end{equation}

Then, the Hessian matrix $\bf H$ can be approximated as ${\bf H} = {\bf J}^T {\bf J}$, and the Levenberg-Marquardt method can be used to compute the gradient. For simplicity, the ``AutoDiff'' method by Ceres Solver~\cite{Agarwal_Ceres_Solver_2022} can be used to compute the gradient. Experiments show that the difference between the two solution methods is negligible.

\subsection{Quadric Representation-Based LiDAR Odometry}
As previously discussed, it is clear that the accuracy of the quadric correspondence process plays an essential role in scan-scan matching. Generally, tens of quadric patches are extracted from a LiDAR scan and a double loop is used to calculate the distance between the quadric patches from the source and the target, in which $d_{i,j} = \sum_{i=1}^{n}\operatorname {d_q}(\boldsymbol c_j, \boldsymbol p_i)$ is the distance between the $i$-th quadric patch in the source and the $j$-th quadric patch in the target. 
\begin{figure}[t]
    \centering
    \subfigure[Mismatched projection of the green point due to the infinite extension of Surface B. For the sake of clarity, 2D points and surfaces are used in this illustration.]{
        \begin{minipage}[t]{0.95\linewidth}
            \centering
            \includegraphics[width=.95\textwidth]{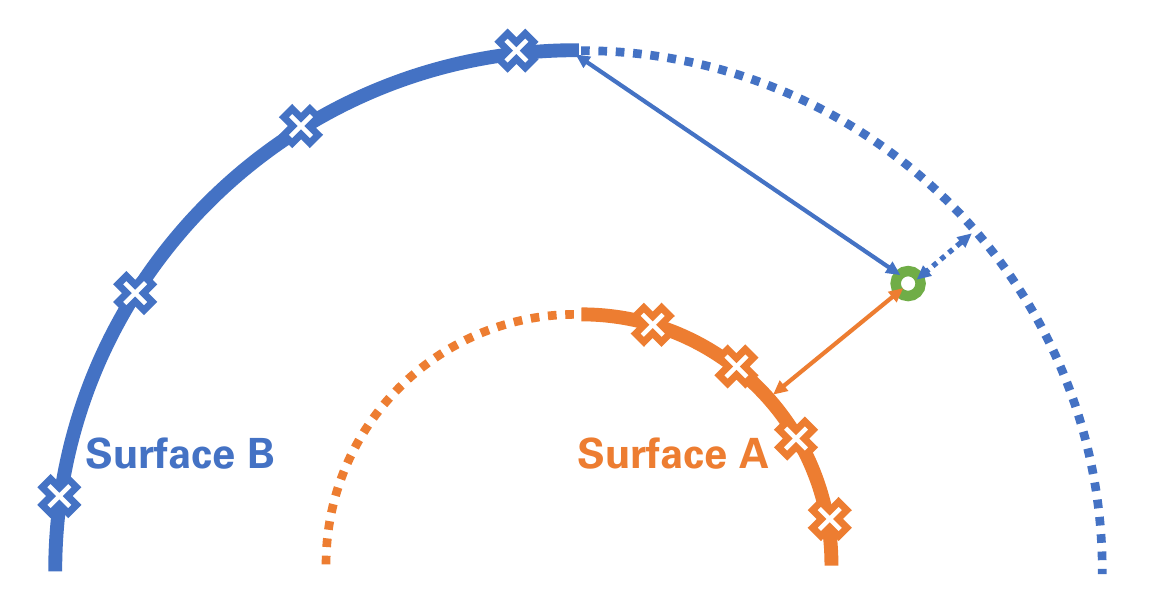}
        \end{minipage}%
        \label{mismatch}
    } \\
    \vspace{-6pt}
    \subfigure[Quadric patch association in real consecutive LiDAR frames.]{
        \begin{minipage}[t]{0.95\linewidth}
            \centering
            \includegraphics[width=.95\textwidth]{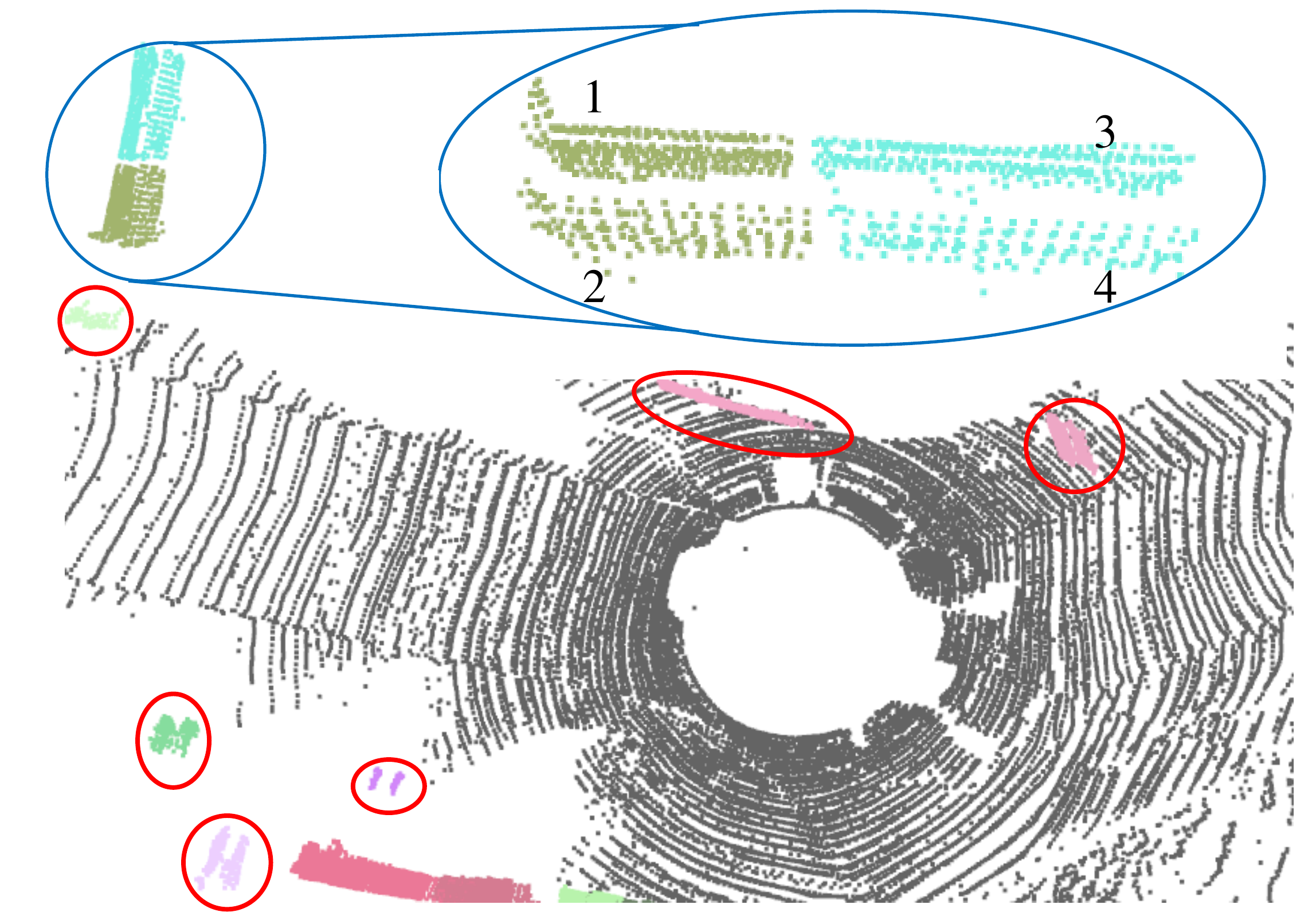}\\
        \end{minipage}%
        \label{association}
    }%
    \caption{Quadric patch association method.}
    \vspace{-6pt}
    \label{correspondence}
\end{figure}

However, since most quadric surfaces are infinite, points may be projected to the incorrect quadric if it is nearest to the extended region of a distantly-centered quadric. As depicted in Fig.~\ref{mismatch}, Surface A and Surface B are quadrics fitted by corresponding color points. The surface may only exist in the vicinity of the points (the solid part of the curve) but not in the dotted part, which is the curve's infinite extension. We see that the new green point from the subsequent LiDAR scan should be projected to Surface A. Unfortunately, the green point is actually closer, in terms of projection distance, to the extended region of Surface B than to Surface A. In this way, mismatched projections can occur due to the infinite extension of most quadrics.

\begin{figure*}[t]
\centering
    \subfigure[First scan.]{\includegraphics[width=.3\textwidth]{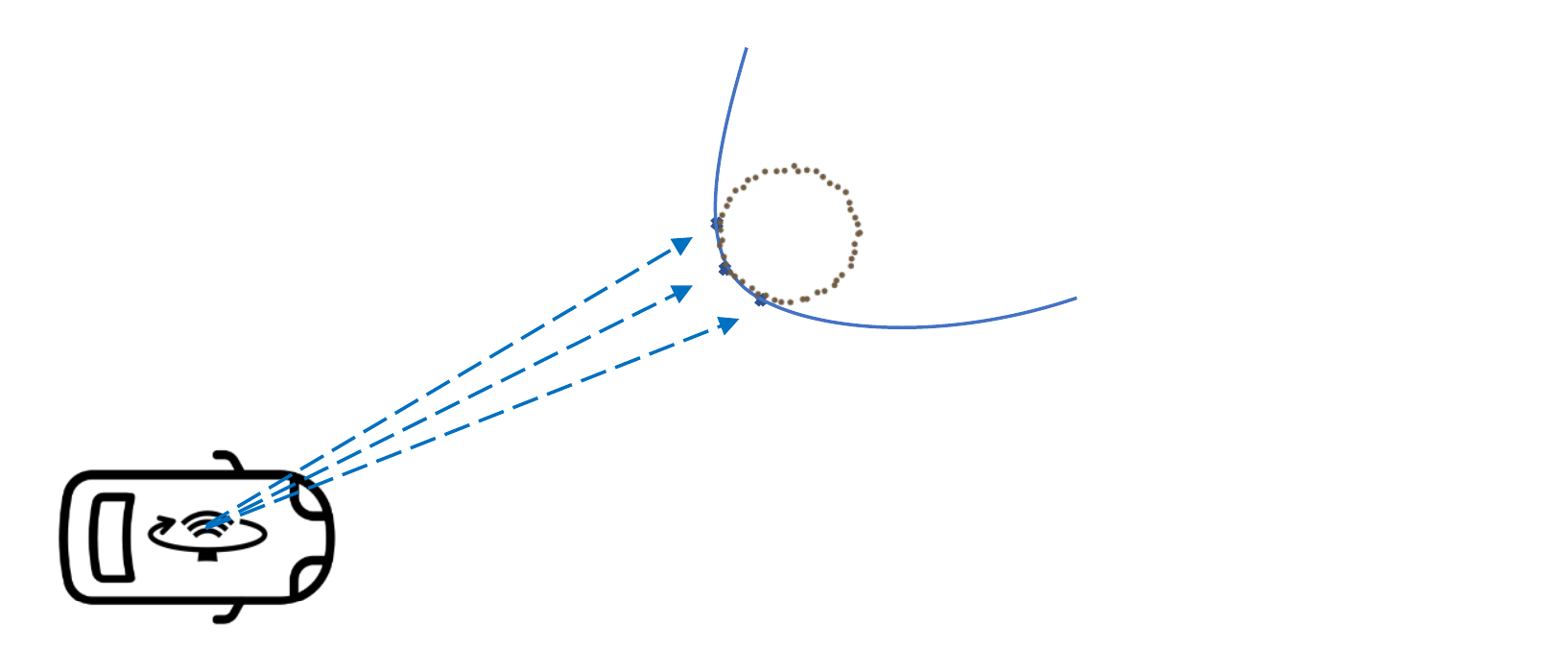}}
    \subfigure[Second scan.]{\includegraphics[width=.3\textwidth]{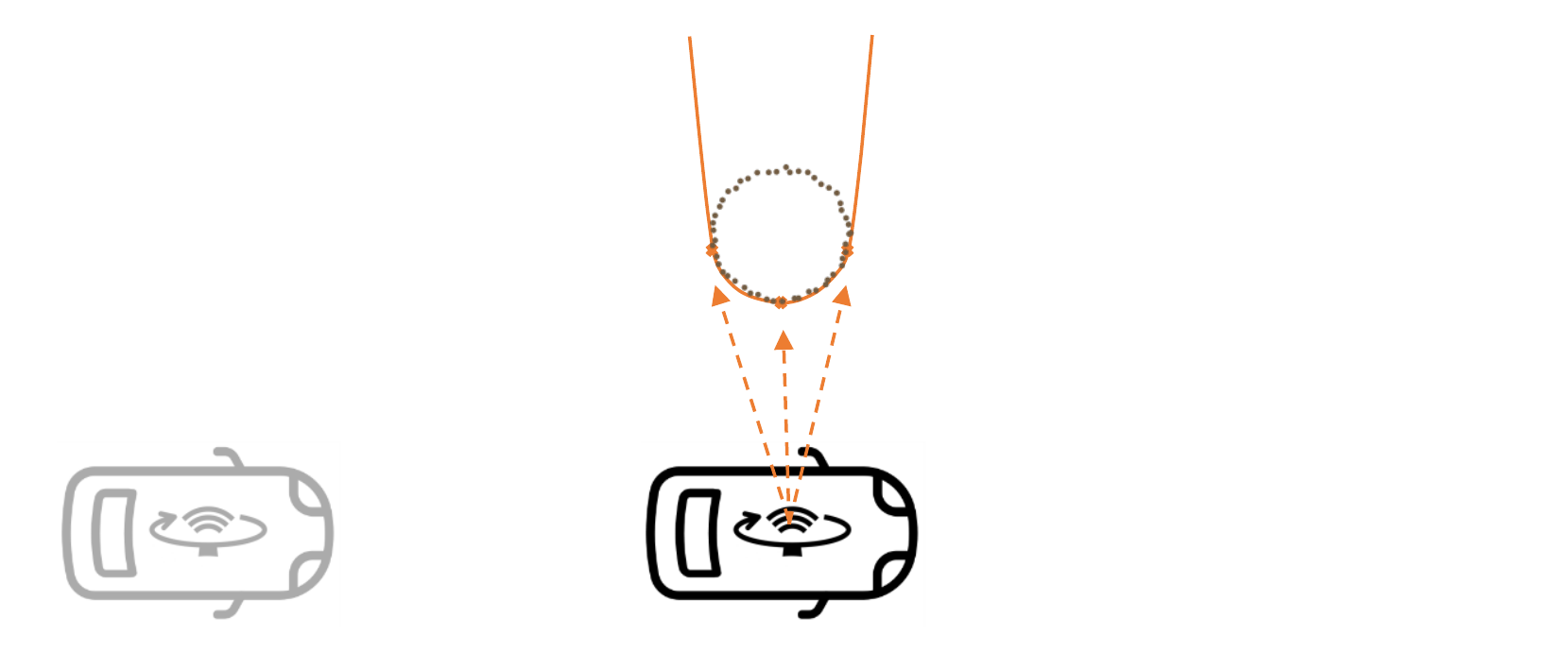}}
    \subfigure[Last scan.]{\includegraphics[width=.3\textwidth]{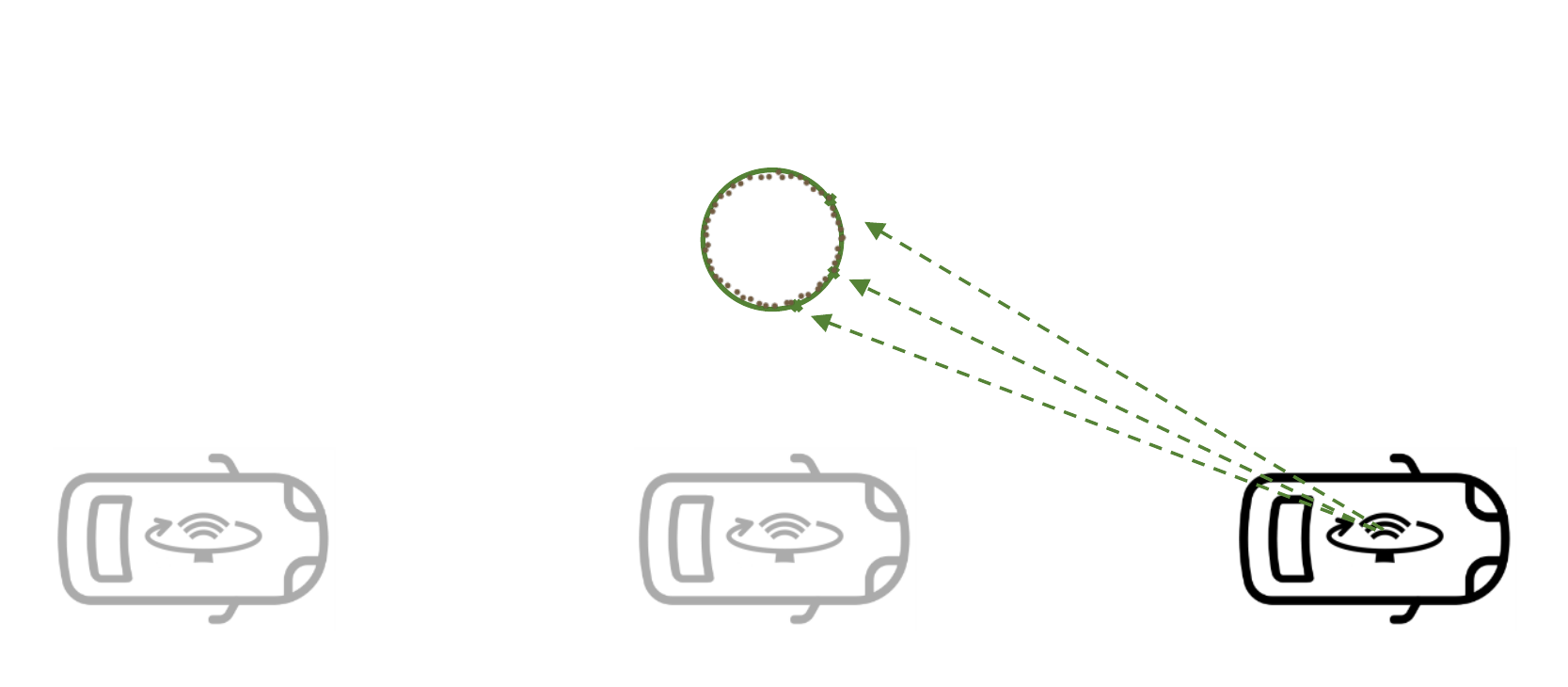}}
    \caption{Quadric representation growing method. For the sake of simplicity, we use a 2D bird-eye view of the 3D point cloud, where the circle represents a static object, such as a tree trunk or pole, which is sampled using consecutive LiDAR scans by the moving vehicle.}
    \label{patch_growing}
\end{figure*}

\subsubsection{Quadric Association}
According to the characteristics of the quadric representation, we design a novel quadric patch association method to drastically reduce the number of mismatched projections. We use a weighted quadric patch-to-patch distance:
\begin{equation}
    d_{weight(i,j)} = \sum_{i=1}^{n} \frac{\alpha \cdot \operatorname {d_q}(\boldsymbol c_j, \boldsymbol p_i)}{\beta + \gamma \cdot \operatorname {exp}(-(\boldsymbol p_i - \boldsymbol \mu_j)^T \mathbf{S}_j^{-1} (\boldsymbol p_i - \boldsymbol \mu_j)}
    \label{weighted}\text{.}
\end{equation}

When the Euclidean distance between two patches is large, the projection distance from the point to the quadric surface increases rapidly, up to $\frac{\alpha}{\beta}$ times the original distance. An example of a quadric patch association result is shown in Fig.~\ref{association}. All the patches in red circles are corresponding patch pairs, whose points are the same color. In addition, two patch pairs (the plane-shaped patched are labeled 1, 2, 3, 4 for clarity) in the blue circle are zoomed in, where we can see that the blue and green walls actually belong to a coincident plane that is split into two parts due to a small gap. Because they are parallel, planes 1 and 3 are fitted to near-identical plane functions, which means that source patches 2 and 4 could correspond to any one of the two target planes. In other words, if the projection distances of points to quadrics are not weighted, then for plane 2 (equivalently, plane 4), the projection distances to 1 and 3 are almost the same, and it is very likely that false associations will occur. If the wrong association happens to two distant patches due to the quadric infinite extension, there will inevitably be registration failure. Our proposed patch association method ensures that planes 2 and 4 are associated with their corresponding planes (planes 1 and 3, respectively) correctly.

\subsubsection{Patch Selection Strategy}
Quadric surfaces exist in a large number of man-made scenes, such as the ground (plane), building surfaces (planes and other quadrics), tree trunks (cylinders), and tree crowns (half-spheres). Indeed, the number and the size of the extracted patches vary drastically in different scenarios. 
Plenty of regular and complete quadric patches can be extracted from urban scenes, but many quadric patches extracted from countryside scenes are fragmented. To maintain the efficiency and robustness of the algorithm, we propose an adaptive-resolution patch selection strategy, where we define a scale factor ($s=\frac{k}{r}$, where $k$ is the point size of a patch and $r$ is the distance from the center location of a patch to the origin) and successively select patches from decreasing scale factors, starting from the largest scale factor, until the point size reaches a threshold.

\subsection{Quadric Representation-Based LiDAR Mapping}
In point-wise LiDAR odometry methods, the feature points extracted from each scan are simply merged into a submap and then downsampled by a voxel grid algorithm, and the NNS algorithm using KD-trees is used to find correspondences with respect to points from a real-time scan. Unfortunately, this process is not suitable for patch-wise methods. On the one hand, reserving all the patches directly from several consecutive frames may lead to redundancy due to patches in different frames describing the same surface. On the other hand, saving all the points to a submap and then re-extracting patches not only increases the runtime of the algorithm, but also causes a decline in the stability of the algorithm. Thus, we propose an incremental map construction method, where corresponding patches from consecutive scans are merged and their implicit functions, center location vectors, and covariance matrices are updated incrementally. 

\subsubsection{Quadric Patch Growing}
As demonstrated in~\cite{vaskevicius2010extraction}, the quadric coefficient vector $\boldsymbol c$ can easily solved using three intermediate matrices $\mathbf{S}$, $\mathbf{P}$, and $\mathbf{Q}$. In fact, all of the parameters in a quadric surface are updated incrementally, including its MSE, point size, and center location. The parameters $\boldsymbol \mu$ and $\overline{\boldsymbol \eta}$ are incremented as follows: 
\begin{subequations}
    \begin{align}
        \boldsymbol{\mu}_{k+1} = \frac{k\boldsymbol{\mu}_k + \boldsymbol{p}_{k+1}}{k+1} \label{incremental1:sub1}\text{,} \\ 
        \overline{\boldsymbol{\eta}}_{k+1} = \frac{k\overline{\boldsymbol{\eta}}_k + \boldsymbol{\eta}_{k+1}}{k+1} \label{incremental1:sub2}\text{.}
    \end{align}
    \label{incremental1}
\end{subequations}
The three intermediate matrices are incremented as follows:
\begin{subequations}
    \begin{align}
    \mathbf{S}_{k+1} &= \mathbf{S}_k+\boldsymbol p_{k+1} \boldsymbol p_{k+1}^T-(k+1) \boldsymbol {\mu}_{k+1} \boldsymbol {\mu}_{k+1}^T+ k\boldsymbol{\mu}_k\boldsymbol {\mu}_{k}^T \label{incremental2:sub1}\text{,}\\
    \mathbf{Q}_{k+1} &= \mathbf{Q}_k+\boldsymbol{\eta}_{k+1} \boldsymbol{\eta}_{k+1}^T-(k+1)  \overline{\boldsymbol \eta}_{k+1} \overline{\boldsymbol \eta}_{k+1}^T+ k\overline{\boldsymbol \eta}_k\overline{\boldsymbol \eta}_{k}^T \label{incremental2:sub2}\text{,} \\
    \mathbf{P}_{k+1} &= \mathbf{P}_k+\boldsymbol {\eta}_{k+1} \boldsymbol p_{k+1}^T-(k+1) \overline{\boldsymbol \eta}_{k+1} \boldsymbol {\mu}_{k+1}^T+ k\overline{\boldsymbol \eta}_k\boldsymbol {\mu}_{k}^T \label{incremental2:sub3}\text{.}
    \end{align}
    \label{incremental2}
\end{subequations}

We can update all of the parameters of the quadric representation of two corresponding patches from a scan-scan matching on the source and target scans using Eq.~\ref{incremental1} and Eq.~\ref{incremental2}. Then, we can re-calculate the implicit function using the updated values of $\mathbf{S}$, $\mathbf{P}$, and $\mathbf{Q}$. After incrementing the parameters for one iteration, new points may be involved in the corresponding quadric planes, such that a valid quadric patch turns into a degenerated quadric patch, and vice versa.
The process of updating a plane's MSE is shown in~\cite{poppinga2008fast}, which we extend to quadrics:
\begin{equation}
    mse_k =\frac{1}{k}(\mathbf{S}'_k+2\mathbf{P}'_k+\mathbf{Q}'_k) + 2\boldsymbol{c}_p^{T}\cdot\boldsymbol{\mu}_k+2\boldsymbol{c}_\eta^{T}\cdot\overline{\boldsymbol{\eta}}_k+c_9^2 
\end{equation}
where $\mathbf{S}'=\boldsymbol{c}_{p}^T\cdot(\mathbf{S}_k+k\boldsymbol \mu_k \boldsymbol \mu_k^T)\cdot\boldsymbol c_{p}$, $\mathbf{P}'=\boldsymbol{c}_{\eta}^T\cdot(\mathbf{P}+k\overline{\boldsymbol \eta}_k\boldsymbol \mu_k^T)\cdot\boldsymbol c_{p}$, and $\mathbf{Q}'=\boldsymbol c_{\eta}^T\cdot(\mathbf{Q}_k+k\overline{\boldsymbol \eta}_k\overline{\boldsymbol \eta}_k^T)\cdot\boldsymbol c_{\eta}$.
For example, the MSE of a degenerated quadric patch may be smaller than the threshold after updating the parameters of the patch growing, such that it becomes a quadric patch with a valid implicit function (no longer a zero vector). Similarly, a quadric patch may become a degenerated quadric patch if its MSE is increased during the incremental updating process. Submap updating is supposed to prune those patches that are far away from the current vehicle position. Compared with point-wise processing methods, our method produces a number of patches that is far smaller than the number of original points. Experiments demonstrate that our method usually produces only tens, and at most hundreds, of patches in a submap. Thus, we can tractably traverse all of the patches and delete ones too far away from the vehicle. We see that as a direct result of using the incremental updating method, the submap updating process can be run in real-time.

An example of the quadric patch growing is shown in Fig.~\ref{patch_growing}, which depicts consecutive LiDAR scans of a tree trunk or pole-like object, which is represented using a circle from a bird-eye view. In the first scan of the object, only a small part of the surface is sampled, and the fitted quadric implicit function is quite different from the object surface shape. As the vehicle moves, more parts of the object are sampled and the quadric implicit function is incrementally updated. We see that in successive scans, the fitted quadric implicit function increasingly resembles the surface of the object. In the last scan, the fitted quadric implicit function exactly resembles the surface of the object. We note that the process of incrementing the parameters of a quadric representation depends only on the parameters of the previous quadric representation and the point cloud in the current frame. It does not depend on the point clouds in any previous frame, which is a key advantage of our proposed incremental updating method.

\subsubsection{Scan-Submap Matching}
The scan-submap matching process very closely resembles the scan-scan matching process. For a new scan, we perform scan-scan matching, project the result to the map coordinate system, and then match it with the submap. Finally, the points from the scan are added to the submap using the proposed incremental updating method. In contrast with the patch selection strategy for LiDAR odometry, the smallest patch threshold is increased to 50 points.

\begin{figure*}[!t]
\centering
    \subfigure[ATE (\%) vs. Latency (ms)]{\includegraphics[width=0.475\linewidth]{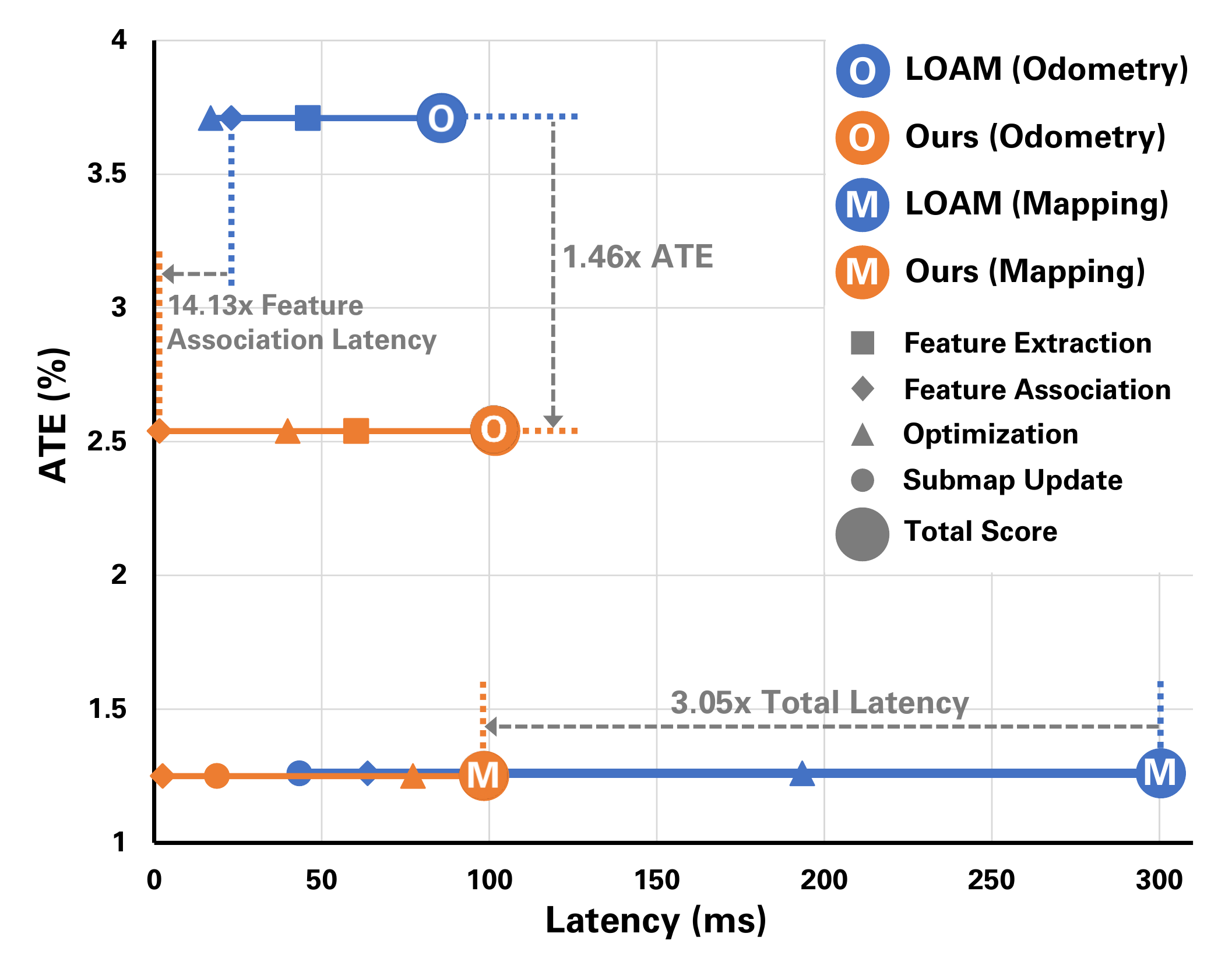}}
    \subfigure[ARE (deg/100m) vs. Latency (ms)]{\includegraphics[width=0.475\linewidth]{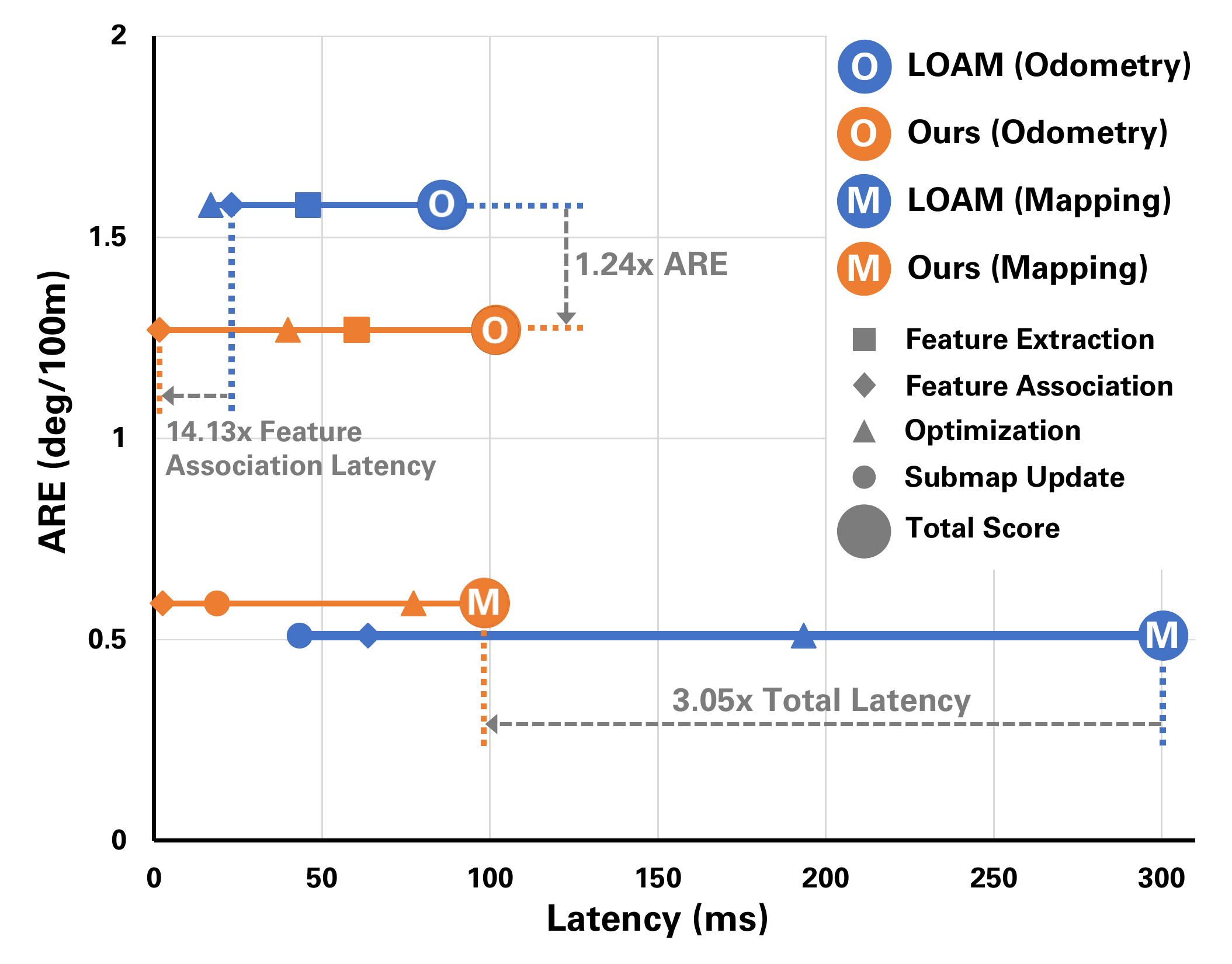}}
    \vspace{-6pt}
    \caption{Comparing both accuracy and latency of our method vs. LOAM for odometry and mapping.}
    \label{time-consuming}
\end{figure*}

\begin{table*}[t]
    \centering
    \tabcolsep 8.8pt
    \caption{ODOMETRY RESULTS ON KITTI 00-10 SEQUENCES}  
    \vspace{-4pt}
    \begin{tabular}{p{1.8cm}<{\centering} | p{0.7cm}<{\centering} | p{0.5cm}<{\centering} p{0.5cm}<{\centering} p{0.5cm}<{\centering} p{0.5cm}<{\centering} p{0.5cm}<{\centering} p{0.5cm}<{\centering} p{0.5cm}<{\centering} p{0.5cm}<{\centering} p{0.5cm}<{\centering} p{0.5cm}<{\centering} p{0.5cm}<{\centering} | p{0.5cm}<{\centering}}
    \toprule 
    Sequence & Metric & 00U & 01H & 02C & 03C & 04C & 05C & 06U & 07U & 08U & 09C & 10C & Avg.\\
    \midrule
    LOAM~\cite{zhang2014loam} & \multirow{2}{*}{ATE} & 4.08 & \textbf{3.31} & 3.43 & 4.63 & \textbf{1.52} & 4.09 & 1.97 & 3.17 & 4.92 & 5.93 & 3.80 & 3.71 \\
    \textbf{OURS} &  & \textbf{2.64} & 5.05 & \textbf{3.31} & \textbf{1.34} & 1.66 & \textbf{1.97} & \textbf{1.30} & \textbf{2.56} & \textbf{2.57} & \textbf{3.32} & \textbf{2.21} & \textbf{2.54}\\
    \midrule
    LOAM~\cite{zhang2014loam} & \multirow{2}{*}{ARE} & 1.69 & \textbf{0.92} & \textbf{1.19} & 2.16 & \textbf{1.14} & 1.68 & 0.82 & \textbf{1.87} & 2.12 & 1.91 & 1.83 & 1.58\\
    \textbf{OURS} &  & \textbf{1.30} & 1.03 & 1.39 & \textbf{1.27} & 1.22 & \textbf{1.01} & \textbf{0.77} & 2.04 & \textbf{1.19} & \textbf{1.40} & \textbf{1.34} & \textbf{1.27}\\
    \bottomrule
    \end{tabular}
    \begin{tablenotes}
        \centering
        \footnotesize
        \item ATE (\%), ARE (deg/100m); U: urban, H: highway, C: countryside. We obtain results for LOAM~\cite{zhang2014loam} using their official source code.
      \end{tablenotes}    
    \label{kitti_odo}
\end{table*}

\begin{table*}[t]
    \centering
    \tabcolsep 8.8pt
    \caption{MAPPING RESULTS ON KITTI 00-10 SEQUENCES}  
    \vspace{-4pt}
    \begin{tabular}{p{1.8cm}<{\centering} | p{0.7cm}<{\centering} | p{0.5cm}<{\centering} p{0.5cm}<{\centering} p{0.5cm}<{\centering} p{0.5cm}<{\centering} p{0.5cm}<{\centering} p{0.5cm}<{\centering} p{0.5cm}<{\centering} p{0.5cm}<{\centering} p{0.5cm}<{\centering} p{0.5cm}<{\centering} p{0.5cm}<{\centering} | p{0.5cm}<{\centering}}
    \toprule 
    Sequence & Metric & 00U & 01H & 02C & 03C & 04C & 05C & 06U & 07U & 08U & 09C & 10C & Avg.\\
    \midrule
    ICP-po2po~\cite{besl1992method} & \multirow{6}{*}{ATE} & 6.88 & 11.21 & 8.21 & 11.07 & 6.64 & 3.97 & 1.95 & 5.17 & 10.04 & 6.93 & 8.91 & 7.36 \\ 
    ICP-po2pl~\cite{besl1992method} & & 3.80 & 13.53 & 9.00 & 2.72 & 2.96 & 2.29 & 1.77 & 1.55 & 4.42 & 3.95 & 6.13 & 4.74 \\ 
    GICP~\cite{segal2009generalized} &  & 1.29 & 4.39 & 2.53 & 1.68 & 3.76 & 1.02 & 0.92 & 0.64 & 1.58 & 1.97 & 1.31 & 1.92 \\ 
    LODONET~\cite{zheng2020lodonet} &  & 1.43 & \textbf{0.96} & 1.46 & 2.12 & 0.65 & 1.07 & 0.62 & 1.86 & 2.04 & \textbf{0.63} & \textbf{1.18} & 1.27\\ 
    LOAM~\cite{zhang2014loam} &  & \textbf{0.83} & 1.96 & 1.75 & \textbf{1.32} & \textbf{0.60} & 0.76 & 0.72 & 1.25 & \textbf{1.47} & 1.42 & 1.78 & 1.26 \\
    \textbf{OURS} &  & 0.92 & 2.61 & \textbf{1.16} & 1.45 & 0.87 & \textbf{0.76} & \textbf{0.55} & \textbf{1.09} & 1.68 & 0.85 & 1.76 & \textbf{1.25}\\
    \midrule
    ICP-po2po~\cite{besl1992method} & \multirow{6}{*}{ARE} & 2.99 & 2.58 & 3.39 & 5.05 & 4.02 & 1.93 & 1.59 & 3.35 & 4.93 & 2.89 & 4.74 & 3.41 \\
    ICP-po2pl~\cite{besl1992method} & & 1.73 & 2.58 & 2.74 & 1.63 & 2.58 & 1.08 & 1.00 & 1.42 & 2.14 & 1.71 & 2.60 & 1.93 \\
    GICP~\cite{segal2009generalized} & & 0.64 & 0.91 & 0.77 & 1.08 & 1.07 & 0.54 & 0.46 & 0.45 & 0.75 & 0.77 & 0.62 & 0.73 \\
    LODONET~\cite{zheng2020lodonet} & & 0.69 & \textbf{0.28} & 0.57 & 0.98 & 0.45 & 0.59 & 0.34 & 1.64 & 0.97 & \textbf{0.35} & \textbf{0.45} & 0.66\\ 
    LOAM~\cite{zhang2014loam} &  & \textbf{0.32} & 0.49 & 0.63 & \textbf{0.64} & \textbf{0.39} & 0.35 & 0.37 & \textbf{0.64} & \textbf{0.54} & 0.60 & 0.69 & \textbf{0.51}\\
    \textbf{OURS} &  & 0.53 & 0.51 & \textbf{0.45} & 0.81 & 0.51 & \textbf{0.33} & \textbf{0.30} & 0.87 & 0.71 & 0.57 & 0.92 & 0.59\\
    \bottomrule
    \end{tabular}
    \begin{tablenotes}
        \centering
        \footnotesize
        \item ATE (\%), ARE (deg/100m); U: urban, H: highway, C: countryside. We obtain results for LOAM~\cite{zhang2014loam} using their official source code.
      \end{tablenotes}    
    \label{kitti_map}
\end{table*}

\subsection{Quadric Representation-Based LiDAR Localization}
The patches generated in the mapping module can also be stored as a global map for localization tasks, as illustrated in the lower right corner of Fig.~\ref{framework}. 
The proposed approach represents the world using quadric representations instead of point clouds, which serves as an efficient way of information extraction and a down-sampling method for the original point cloud scene.
Compared ``dense'' point cloud, this ``sparse'' representation method significantly reduces the memory requirement as large-volume point clouds are avoided.
For localization tasks, both valid and degenerated quadric patches can be loaded from a saved checkpoint to re-create the global map. Following the GeoHash-based encoding method proposed in \cite{xia2022onboard}, the center location of each quadric representation can be encoded into a binary code, 
allowing for easy retrieval of the quadric representation through a double hash function with $O(1)$ time complexity. 
This reduces the time consumption for NNS on a global navigation map by only searching neighboring quadric patches within a certain radius of the vehicle's current position.

\section{EXPERIMENTS}
\noindent \textbf{Datasets.}
We evaluate the proposed quadric representation for LiDAR odometry, mapping and localization both qualitatively and quantitatively on the KITTI~\cite{Geiger2012CVPR} (Velodyne64E) and UrbanLoco~\cite{wen2020urbanloco} (RoboSense32) datasets. 
\textbf{1)} KITTI is a commonly used dataset for autonomous driving, containing various scenes such as urban, rural, and highways. We conduct experiments on the odometry subset (KITTI 00-10 sequences), which provides standard data and evaluation metrics for visual LiDAR SLAM. 

\textbf{2)} UrbanLoco is a more challenging dataset consisting of data collected in highly urbanized areas with a full sensor suite. It covers a wide range of urban terrains such as urban canyons, bridges, tunnels, and sharp turns.
In this work, we select 4 sequences (CA-20190828155828, CA-20190828173350, CA-20190828184706, and CA-20190828190411) from the UrbanLoco dataset collected in San Francisco, and omit the other 3 sequences with challenging long and straight tunnel scenarios that are not suitable for LiDAR odometry, mapping and localization tasks. 

\smallskip

\noindent \textbf{Implementation details.}
In our experiments, we set $\alpha = 1.0, \beta = 0.1, \gamma = 1.9$ (in Eq.~\ref{weighted}), and $mse = 0.04$. All experiments are conducted on a PC equipped with an Intel Core i7-7820HQ@2.90GHz$\times$8 core CPU and 32 GB memory.
To measure the accuracy of our method, we use average translation error (ATE, \%) and average rotation error (ARE, deg/100m) as defined in~\citep{Geiger2012CVPR}. Latency (ms) is used to evaluate time consumption.

As a baseline for comparison, we choose LOAM~\cite{zhang2014loam}, a state-of-the-art representation-based method for LiDAR odometry and mapping that uses the same evaluation standards as we do for our approach. Additionally, while our main focus is not on learning-based methods, we include a comparison with LODONET~\cite{zheng2020lodonet} for comprehensiveness. 
More details can be found in the supplementary material.

\subsection{LiDAR Odometry and Mapping on KITTI}
We conduct empirical experiments on the KITTI odometry set and provide the time cost analysis of our quadric representation against previous works that use point representations. We use average translation error (ATE) and average rotation error (ARE) as the evaluation metrics.

\smallskip

\begin{figure}[t]
    {\centering
    \includegraphics[width=0.50\linewidth]{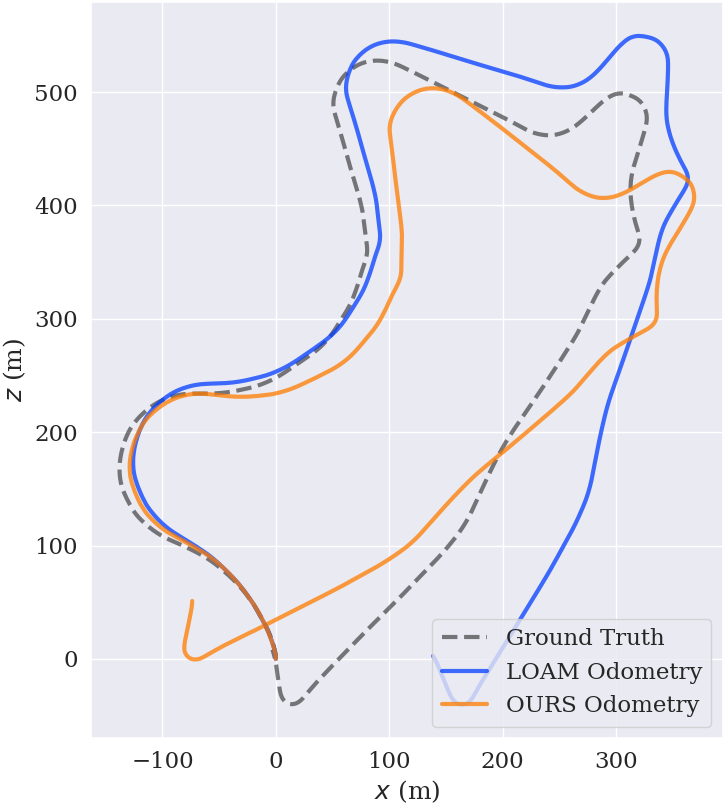}~
    \includegraphics[width=0.485\linewidth]{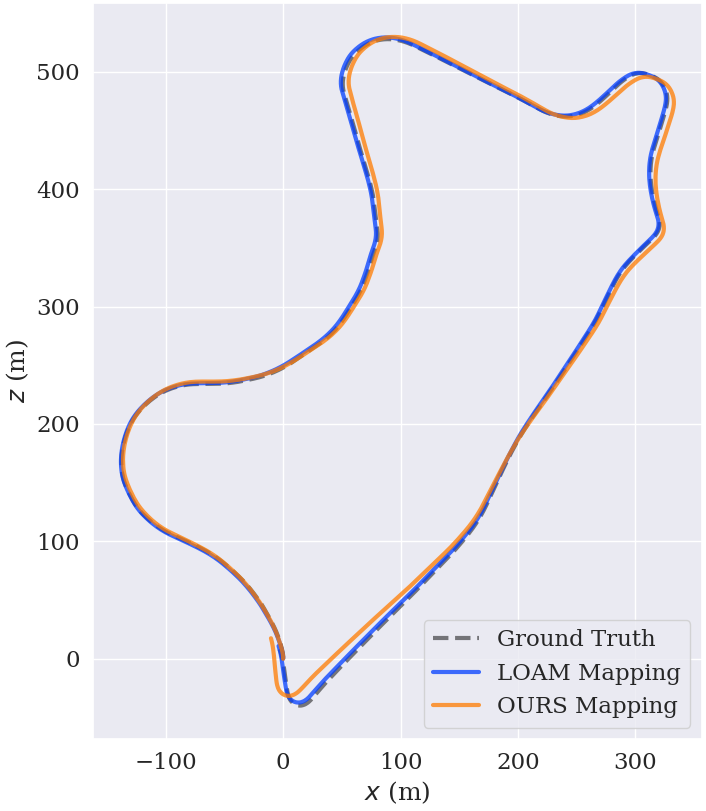}}
    \scriptsize{
    \hspace*{10pt}(a) Odometry trajectories on KITTI 09 \hspace{13pt}
    (b) Mapping trajectories on KITTI 09
    }
    \caption{Comparison results of LiDAR odometry and mapping on the KITTI 09 sequence.}
    \vspace{-6pt}
    \label{kitti09}
\end{figure}

\noindent \textbf{Odometry.}
We first evaluate the proposed quadric representation on LiDAR odometry with our scan-scan matching framework. The experimental results are presented in Tab.~\ref{kitti_odo}, showing the superior results of our method when compared to the previous state-of-the-art method, LOAM~\cite{zhang2014loam}.
For example, the average ATE and ARE (over 11 sequences) of our work are 2.54 and 1.27, respectively, presenting an improvement of 46.1\% and 24.4\% compared to the previous SOTA (3.71 and 1.58).
An example of a qualitative odometry trajectory~\cite{grupp2017evo} (KITTI sequence 09) comparison of our method and LOAM is shown in Fig.~\ref{kitti09} (a). 

The improved performance can be attributed to the following factors.
\textbf{1)} We eliminate possible mismatches and reduce the scan-scan registration error by leveraging quadric representation association in the projection of points to surfaces.
\textbf{2)} To reduce the over-fitting caused by the excessive ground size, we further divide the extracted ground points into 4-6 parts along the x-axis evenly, which proved to be effective for z-axis accuracy.
\textbf{3)} We follow the idea of two-step optimization~\cite{shan2018lego}, where ground quadric representations are used for the estimation of $[pitch, roll, z]$ while the rest are used for the estimation of $[yaw, x, y]$. 

\smallskip

\noindent \textbf{Mapping.}
We then evaluate the mapping performance of our proposed quadric representation on KITTI with both scan-scan matching and scan-submap correction modules. We compare our method with point-based works including LODONET~\cite{zheng2020lodonet} and LOAM~\cite{zhang2014loam}.
As shown in Tab.~\ref{kitti_map}, our method outperforms them in ATE and achieves competitive performance in ARE, demonstrating its effectiveness in LiDAR mapping.
The example of mapping trajectory comparison on KITTI sequence 09 is shown in Fig.~\ref{kitti09} (b).

Note that, typically, around 12000 planar points and 5000 edge points are selected from a scan, along with around 50000 planar points and 15000 edge points saved in a submap for mapping fine-tuning.
This ``dense'' point cloud registration method matches real-time scan and submap with sufficient accuracy but is calculation-intensive and time-consuming.
This results in a low mapping frequency in LOAM of only 1 Hz compared to the 10 Hz in the odometry module.
However, benefiting from the proposed submap generation method
which uses quadric representations instead of points, our work offers a simpler and more time-efficient paradigm.
We provide a detailed time efficiency comparison in Fig.~\ref{time-consuming}.

\begin{figure*}[t]
    \centering
    \includegraphics[height=9em]{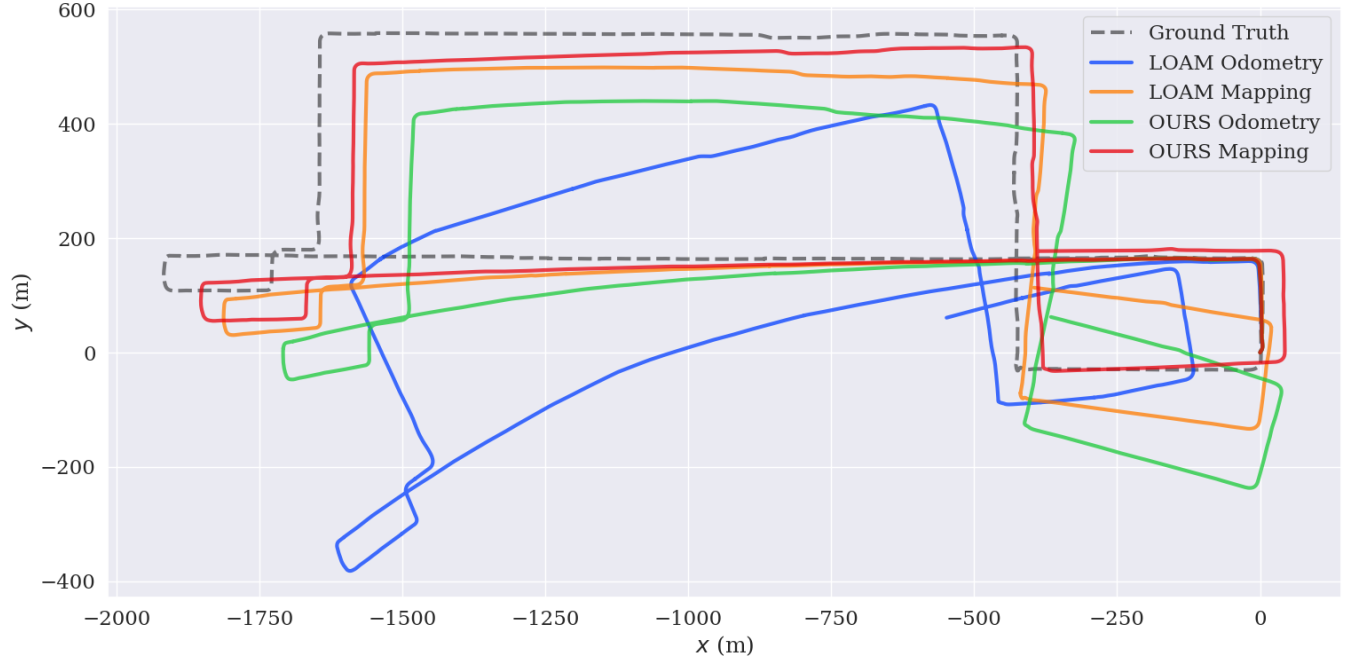}
    \includegraphics[height=9em]{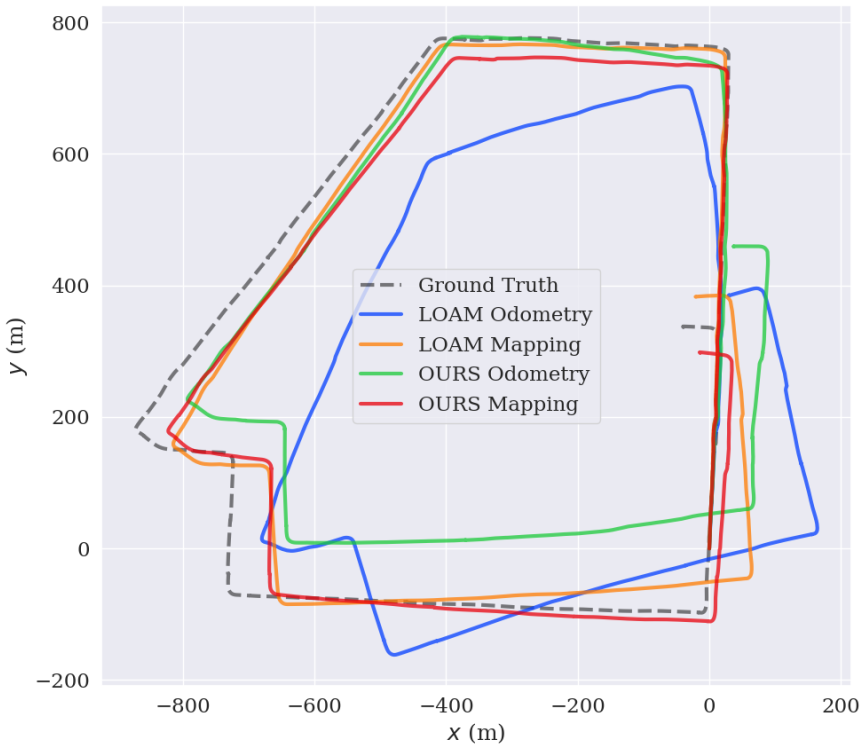}
    \includegraphics[height=9em]{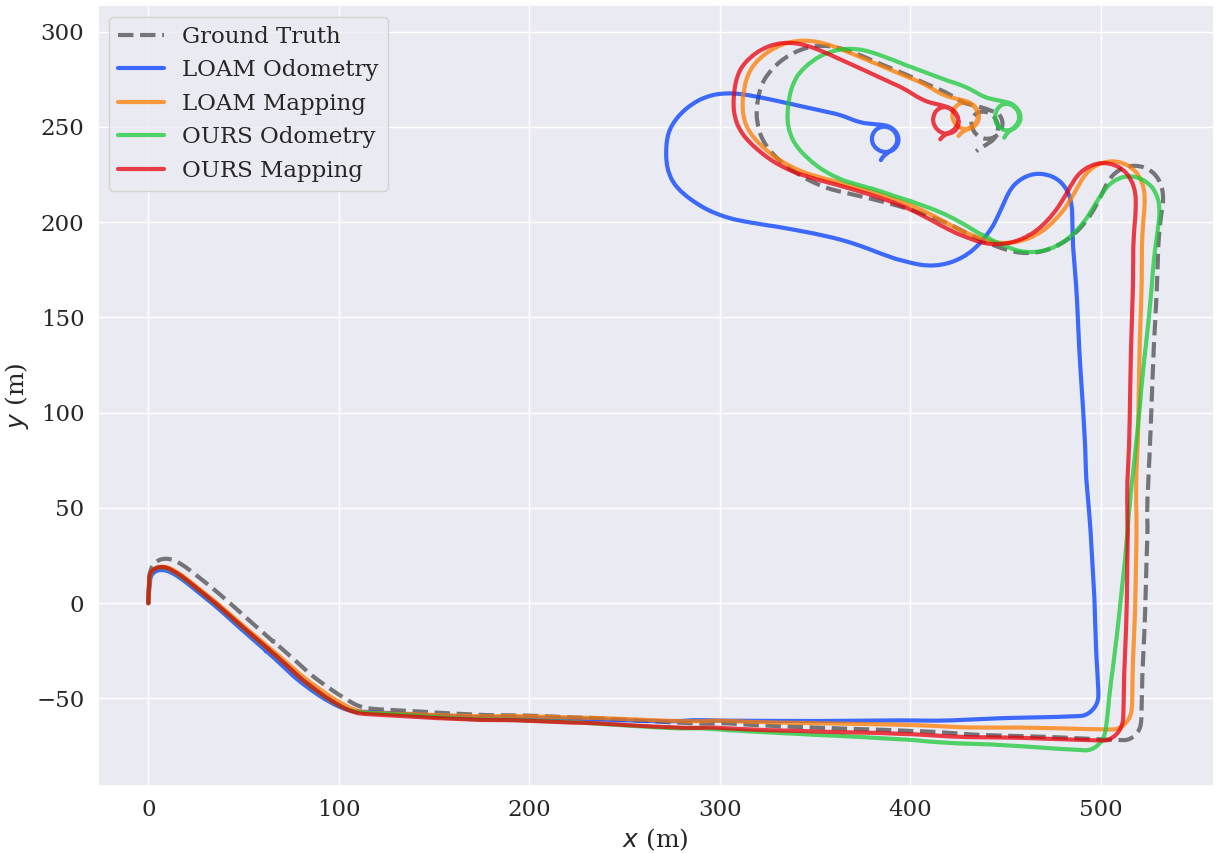}
    \includegraphics[height=9em]{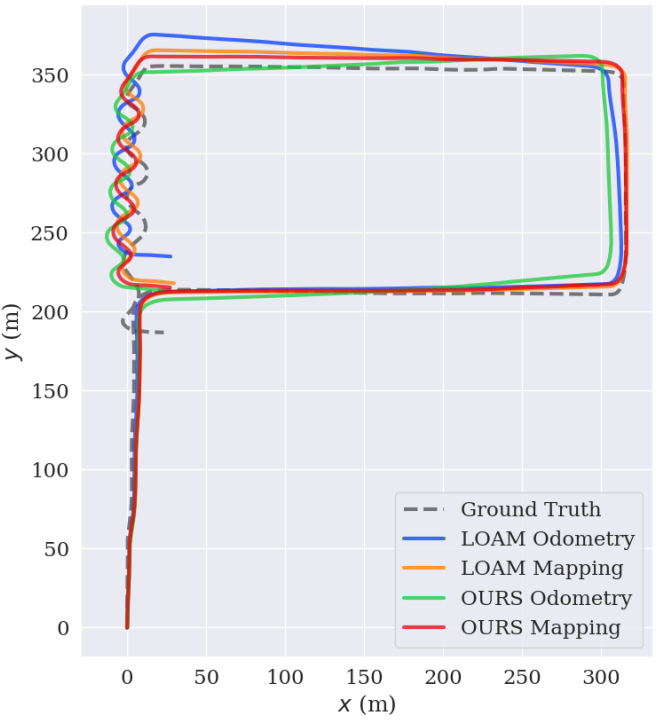}
    \scriptsize{
    \hspace*{65pt} (a) CA-20190828155828 \hspace{30pt} \hfill
    (b) CA-20190828173550 \hspace{15pt} \hfill
    (c) CA-20190828184706 \hfill
    (d) CA-20190828190411} \hspace{2pt}
    \caption{Comparison results of LiDAR odometry and mapping on UrbanLoco sub-dataset.}
    \label{urbanloco}
\end{figure*}

\begin{table*}
    \small 
    \centering
    \scriptsize 
    \caption{COMPARISON RESULTS OF ODOMETRY ON URBANLOCO ROSBAGS}
    \begin{tabular}{c | c | p{1.5cm}<{\centering} | p{1.0cm}<{\centering}  p{1.0cm}<{\centering}  p{1.0cm}<{\centering} p{1.0cm}<{\centering} | p{1.0cm}<{\centering}  p{1.0cm}<{\centering}  p{1.0cm}<{\centering}  p{1.0cm}<{\centering}}
        \toprule
        \multirow{2.5}{*}{ROSBAG} & \multirow{2.5}{*}{Length (km)} & \multirow{2.5}{*}{Method} & \multicolumn{4}{c|}{Mean Translation Error (m)} & \multicolumn{4}{c}{Mean Rotation Error (degree)} \\
        \cmidrule{4-11}
         & & & X & Y & Z & Ave. & Yaw & Pitch & Roll & Ave. \\
         \midrule
         \multirow{2}{*}{CA-20190828155828} & \multirow{2}{*}{5.9} & LOAM & 123.84 & 103.79 & 340.17 & 189.27 & 12.89 & \textbf{8.73} & 5.95 & \textbf{9.19} \\
         & & \textbf{OURS} & \textbf{58.67} & \textbf{102.46} & \textbf{239.43} & \textbf{133.52} & \textbf{10.99} & 13.84 & \textbf{4.58} & 9.80 \\
         \midrule
         \multirow{2}{*}{CA-20190828173350} & \multirow{2}{*}{3.2}
         & LOAM & \textbf{69.38} & 98.57 & \textbf{179.89} & \textbf{99.28} & 23.93 & \textbf{15.89} & \textbf{19.94} & 19.92 \\
         & & \textbf{OURS} & 76.38 & \textbf{62.64} & 191.93 & 106.98 & \textbf{10.54} & 20.12 & 22.35 & \textbf{17.67}  \\
         \midrule
         \multirow{2}{*}{CA-20190828184706} & \multirow{2}{*}{1.8}
         & LOAM & 50.01 & 14.61 & 87.35 & 50.69 & 8.17 & \textbf{6.90} & 9.30 & 8.12 \\
         & & \textbf{OURS} & \textbf{14.72} & \textbf{7.13} & \textbf{63.32} & \textbf{28.39} & \textbf{6.64} & 9.29 & \textbf{7.10} & \textbf{7.68} \\
         \midrule
         \multirow{2}{*}{CA-20190828190411} & \multirow{2}{*}{1.0}
         & LOAM & \textbf{4.34} & 48.16 & \textbf{11.10} & \textbf{21.20} & \textbf{4.54} & 10.24 & \textbf{1.38} & \textbf{5.44} \\
         & & \textbf{OURS} & 5.83 & \textbf{38.42} & 35.24 & 26.50 & 4.83 & \textbf{6.05} & 5.61 & 5.49 \\
        \bottomrule
    \end{tabular}
    \label{urbanloco_odo}
\end{table*}

\begin{table*}
    \small 
    \centering
    \scriptsize 
    \caption{COMPARISON RESULTS OF MAPPING ON URBANLOCO ROSBAGS}
    \begin{tabular}{c | c | p{1.5cm}<{\centering} | p{1.0cm}<{\centering}  p{1.0cm}<{\centering}  p{1.0cm}<{\centering} p{1.0cm}<{\centering} | p{1.0cm}<{\centering}  p{1.0cm}<{\centering}  p{1.0cm}<{\centering}  p{1.0cm}<{\centering}}
        \toprule
        \multirow{2.5}{*}{ROSBAG} & \multirow{2.5}{*}{Length (km)} & \multirow{2.5}{*}{Method} & \multicolumn{4}{c|}{Mean Translation Error (m)} & \multicolumn{4}{c}{Mean Rotation Error (degree)} \\
        \cmidrule{4-11}
         & & & X & Y & Z & Ave. & Yaw & Pitch & Roll & Ave. \\
         \midrule
         \multirow{2}{*}{CA-20190828155828} & \multirow{2}{*}{5.9} & LOAM & \textbf{29.24} & 51.23 & 119.71 & 66.73 & 7.39 & 8.55 & 4.58 & 6.84 \\
         & & \textbf{OURS} & 36.57 & \textbf{12.69} & \textbf{98.48} & \textbf{49.24} & \textbf{5.63} & \textbf{6.92} & \textbf{2.13} & \textbf{4.89} \\
         \midrule
         \multirow{2}{*}{CA-20190828173350} & \multirow{2}{*}{3.2}
         & LOAM & \textbf{18.13} & 45.47 & 90.02 & 51.21 & 4.37 & 1.05 & 10.50 & 5.31 \\
         & & \textbf{OURS} & 25.37 & \textbf{39.01} & \textbf{79.82} & \textbf{48.01} & \textbf{4.08} & \textbf{1.02} & \textbf{6.60} & \textbf{3.9}  \\
         \midrule
         \multirow{2}{*}{CA-20190828184706} & \multirow{2}{*}{1.8}
         & LOAM & \textbf{9.30} & \textbf{8.01} & 46.31 & 21.21 & \textbf{2.40} & 1.87 & \textbf{1.27} & \textbf{1.85} \\
         & & \textbf{OURS} & 18.83 &  \textbf{6.29} & \textbf{35.25} & \textbf{20.12} & 2.73 & \textbf{1.78} & 1.66 & 2.06 \\
         \midrule
         \multirow{2}{*}{CA-20190828190411} & \multirow{2}{*}{1.0}
         & LOAM & 6.53 & 31.26 & \textbf{23.85} & 20.55 & 4.34 & \textbf{3.53} & \textbf{0.86} & \textbf{2.91} \\
         & & \textbf{OURS} & \textbf{3.94} & \textbf{28.67} & 28.50 & \textbf{20.37} & \textbf{4.30} & 5.95 & 4.47 & 4.91 \\
        \bottomrule
    \end{tabular}
    \label{urbanloco_map}
\end{table*}

\smallskip

\noindent \textbf{Time cost analysis.} 
We have divided the odometry module and mapping module into three distinct sub-modules. The first sub-module of odometry focuses on feature extraction, the second sub-module deals with associating selected feature points with the target scan, and the third sub-module performs pose estimation. On the other hand, the mapping sub-module consists of associating feature points with the submap, estimating the pose, and updating the submap. Most LiDAR odometry algorithms are composed of these sub-modules. 

Point-wise registration methods, like LOAM odometry methods, can often achieve a high frequency in the scan-scan matching process due to the relatively low point cloud density in a single frame.

However, when considering the mapping module, the direct impact of the increase of point cloud size in the submap is the decrease in frame rate. Experiments show that the average time consumption of scan-submap matching for LOAM is around 300 ms,
whereas our method's time is around 100 ms.

The detailed time consumption of each module comparison between LOAM and the proposed algorithm is shown in Fig.~\ref{time-consuming}. 

The number of selected feature points successfully associated with a target quadric representation by our method is around 5000 to 10000 points in the KITTI odometry dataset. This is far larger than the number selected by LOAM, which is no more than 1224 planar points and 612 edge points. We see that for odometry, our proposed method consumes more time for feature extraction than LOAM. However, the advantage of our patch/surface-wise method is reflected in the quadric representation association sub-module. We see that our proposed method is very efficient in the quadric-quadric association process, as the number of quadric representations is much smaller than in point clouds. 
This advantage is further amplified in scan-submap registration, where the average time consumption for feature association is increased from 1.68 ms to 2.56 ms for our method, but is increased to 63.69 ms in LOAM scan-submap registration. 
In addition, our method is more efficient than LOAM's point-wise operation in terms of submap updating, as it can directly add point clouds from the current scan to their corresponding quadric representation incrementally after the scan-submap association sub-module.

\begin{table*}[t]
    \centering
    \caption{COMPARISON OF MAP REPRESENTATION METHODS FOR NNS AND POSE ESTIMATION ACCURACY}
    \begin{tabular}{c c | c c c | c c c c}
    \toprule
         Element & Occupation & NNS Method & complexity & $L_{nns}$ (ms) & Registration Method & $E_t$ (cm) & $E_r$ (degree) & $L_{reg}$ (ms) \\
         \midrule
         Point Cloud & 16.9 GB & KD-TREE & $O(log(n))$ & 290 & GICP~\cite{segal2009generalized} & 7.90 & 0.051 & 691\\ 
         \midrule
         Voxel & 29.4 MB & Traverse & $O(n)$ & 13 & NDT~\cite{ndt} & 8.29 & 0.049 & 189 \\
         \midrule
         Quadric Representation & \textbf{11.2 MB} & Double Hash & $O(1)$ & \textbf{1.2} & OURS & \textbf{7.43} & \textbf{0.042} & \textbf{94} \\
         \bottomrule
    \end{tabular}
    \begin{tablenotes}
        \footnotesize
        \centering
        \item $L_{nns}$: Average time consumption for once NNS (ms). $L_{reg}$: Average time consumption for once registration.\\
        $E_t$: Translation Root Mean Square Error. $E_r$: Orientation Root Mean Square Error.
      \end{tablenotes} 
    \label{globalmap}
\end{table*} 

\subsection{LiDAR Odometry and Mapping on UrbanLoco}
We also evaluate our proposed quadric representation-based method for LiDAR odometry and mapping on the UrbanLoco Dataset. The ground truth is provided by Novatel SPAN-CPT, a GNSS-IMU navigation system.
Since the frequency of LiDAR and navigation system messages are different, here we simply synchronize them. We first convert the ground truth (``/novatel\_data/inspvax'', 20 HZ) message from WGS84 to the UTM coordinate system and assume that the vehicle maintains constant motion between two consecutive Novatel messages (about a period of 50 ms). Thus we obtained the ground truth pose information of each LiDAR frame (``/rslidar\_points'', 10 HZ) according to the time stamps by performing linear interpolation between the preceding and succeeding Novatel frames.
Mean Translation Error (MTE) and Mean Rotation Error (MRE) are used as evaluation metrics for this dataset.

\smallskip

\noindent \textbf{Odometry.}
We present the LiDAR odometry results of our method and LOAM on UrbanLoco in Tab.~\ref{urbanloco_odo}.
Due to the more complex and dynamic environment captured in the UrbanLoco dataset, the LiDAR odometry accuracy decreases as compared to KITTI odometry accuracy.
We observe that, in this more challenging setting, our quadric representation-based method is still superior to its point-based counterpart.
For example, our method outperforms LOAM by large margins on (CA-20190828)155828 and 184706, especially in terms of MTE (e.g. 58.67 vs. 123.84, and 14.72 vs. 50.01 on the x-axis).

On the other two sequences, we observe comparable performance.

\noindent \textbf{Mapping.}
We evaluate the LiDAR mapping results on UrbanLoco with the same setting as the LiDAR odometry.
As shown in Tab.~\ref{urbanloco_map}, our method consistently shows large improvements over LOAM on all four sequences in terms of MTE, demonstrating the effectiveness of our proposed quadric representations.
As for MTE, our method shows its superiority on (CA-20190828)155828 and 173350, which are the two longest sequences, indicating the proposed quadric representation works better on longer trajectories. 

The trajectory comparisons for both odometry and mapping on all four sequences are presented in Fig.~\ref{urbanloco}. Our method demonstrates consistent improvements over LOAM. 
This can be attributed to the advantage of quadric representations in complex and dynamic environments: In the presence of the complex plane and surface features of man-made constructs in highly-urbanized scenes, our quadric representation-based method is able to maintain high performance.

Compared with Velodyne64E, RoboSense32 LiDAR obtains sparser and rougher local feature information for each point due to fewer scan lines, which is not friendly to the point-wise method. Coupled with the increase of dynamic objects, the registration error will increase as well. However, our method finds surfaces more consistently on a larger scale than the point-wise method, and surface information will not be blurred due to the sparseness of the sampled points.
In this way, our proposed representation demonstrates stronger generalization abilities than previous works.

\subsection{Map Compression and LiDAR Localization}
We have demonstrated that the proposed quadric representation provides accurate LiDAR odometry and mapping results while being more efficient compared to state-of-the-art methods, which suggests this representation method is capable of providing sufficient information for pose estimation.
For map compression and LiDAR localization, as the scale of the scene increases, the marginal efficiency advantage of the proposed quadric representations over point representations will increase.
This is because feature association in LiDAR localization becomes increasingly challenging due to the high volume of global navigation maps.
Thus, quadric representations boast a substantial advantage over point representations for storage efficiency and correspondence searching. 
We also leverage a GeoHash-based encoding technique~\cite{xia2022onboard} 
for fast correspondence searching, where the quadric representation center location is encoded to a binary code (encoding resolution is set to $1.5m\times1.5m\times1.5m, (x-y-z)$). In order to ensure the uniqueness of each binary code, only one quadric with the smallest $mse$ is reserved.

We follow \cite{xia2022onboard} to conduct the localization experiment. We add random noise on the basis of the ground truth pose to generate the initial pose for scan-map registration. The noise is subjected to a uniform distribution: $\mathcal{U}(-0.2,0.2)$ (m) in the x and y axis, while the noise in the yaw direction is: $\mathcal{U}(-5.0,5.0)$ (degree).
As shown in Tab.~\ref{globalmap}, our quadric representation occupies 11.2MB of storage, saving 2.6$\times$ and 1508.9$\times$ respectively compared to voxel and point cloud representations.
Moreover, the computational complexity of our method for correspondence searching is only $O(1)$, providing a significant advantage over voxel ($O(log(n))$) and point cloud ($O(n)$) based methods.
The quantitative results in Tab.~\ref{globalmap} show that our scan-map registration outperforms two classical point-wise (GICP) and voxel-wise (NDT) registration algorithms.

\section{CONCLUSION} 
We propose a quadric representation-based method for LiDAR odometry, mapping and localization tasks. In our method, patches are extracted from 3D scenes and then fitted to different types of quadrics, after which point-to-quadric representation registration is utilized for LiDAR pose estimation. We also present a novel incremental growing method for quadric representations, which eliminates the need for the time-consuming process of re-fitting quadric surfaces. Experiments on KITTI and UrbanLoco datasets demonstrate that our quadric representation-based method has a significantly better time-efficiency than point-wise methods (e.g. 3.05$\times$ LOAM for mapping total latency), while achieving superior accuracy on odometry tasks and competitive accuracy on mapping tasks (e.g. 1.64$\times$ LOAM for odometry ATE). Furthermore, quadric representations, encoded by a GeoHash-based method, can be saved for future localization tasks to avoid repeated iterations of NNS, which is the most time-consuming module for localization in large-scale scenarios.

\newpage

\twocolumn[{
\renewcommand\twocolumn[1][]{#1}%
{\centerline{\large \bf APPENDIX}}
\vspace{20pt}
\begin{center}
\includegraphics[height=1.07in]{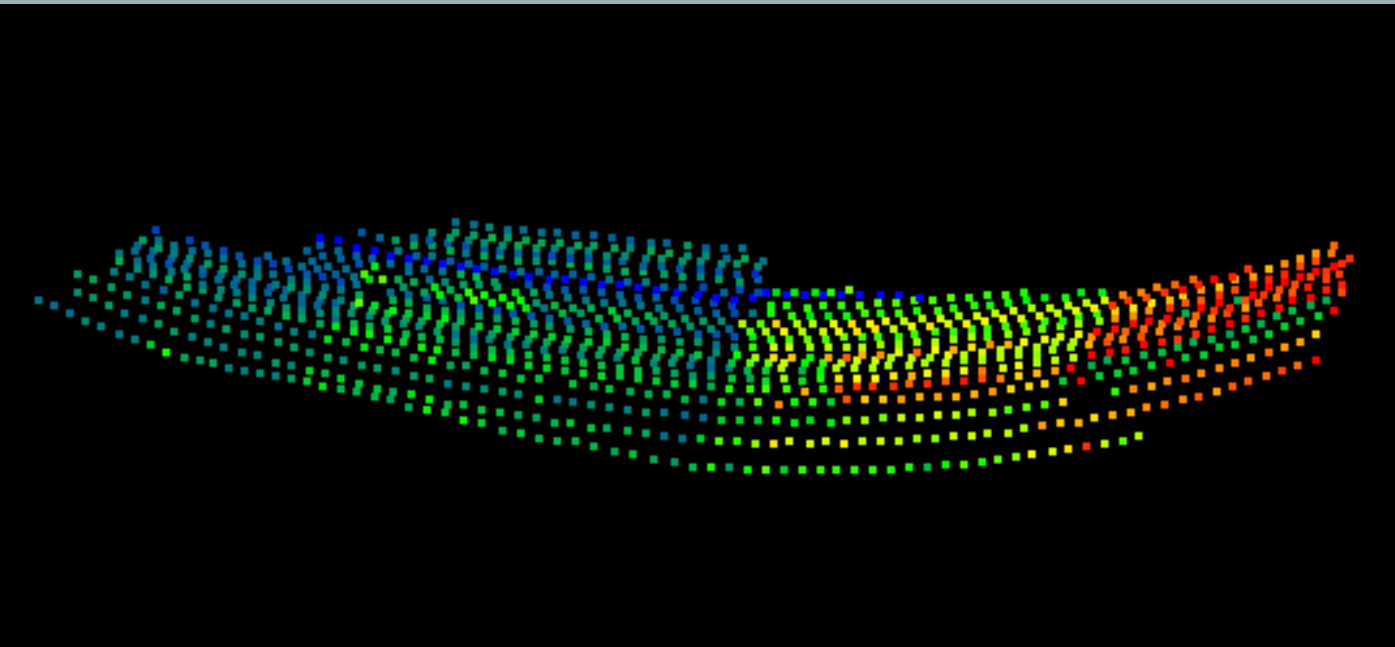}
\includegraphics[height=1.07in]{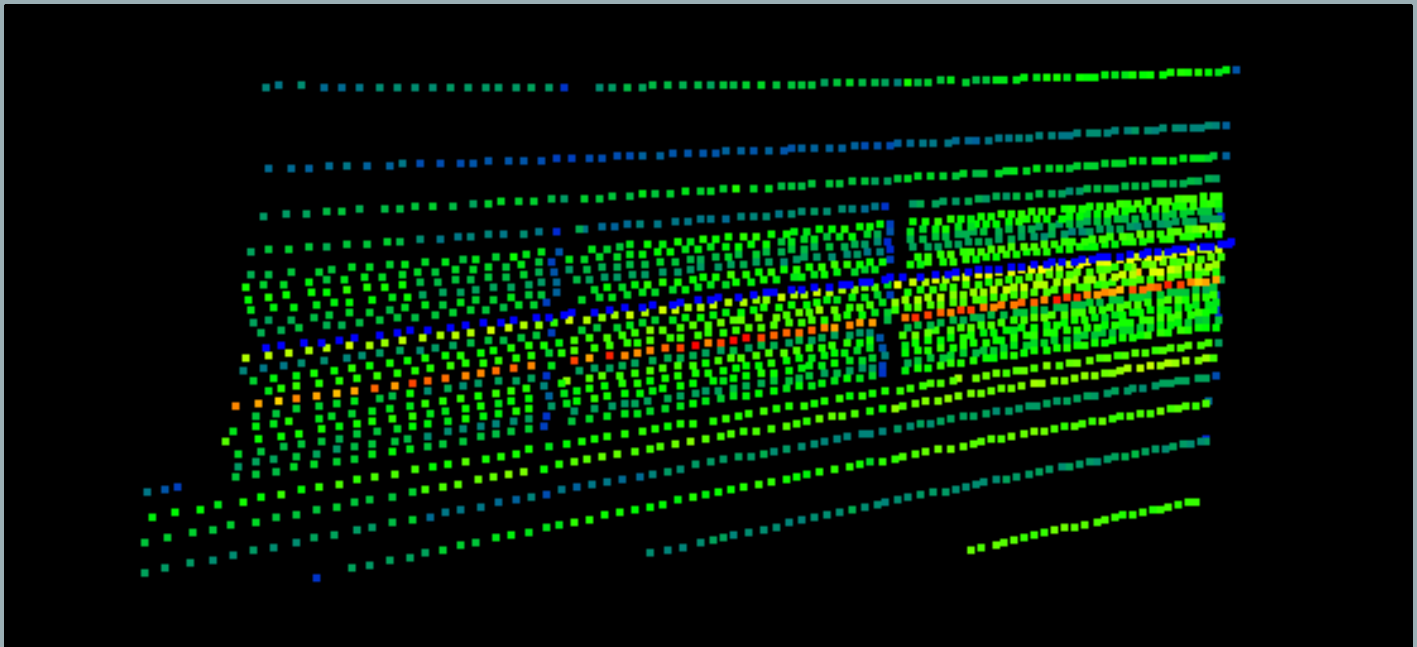}
\includegraphics[height=1.07in]{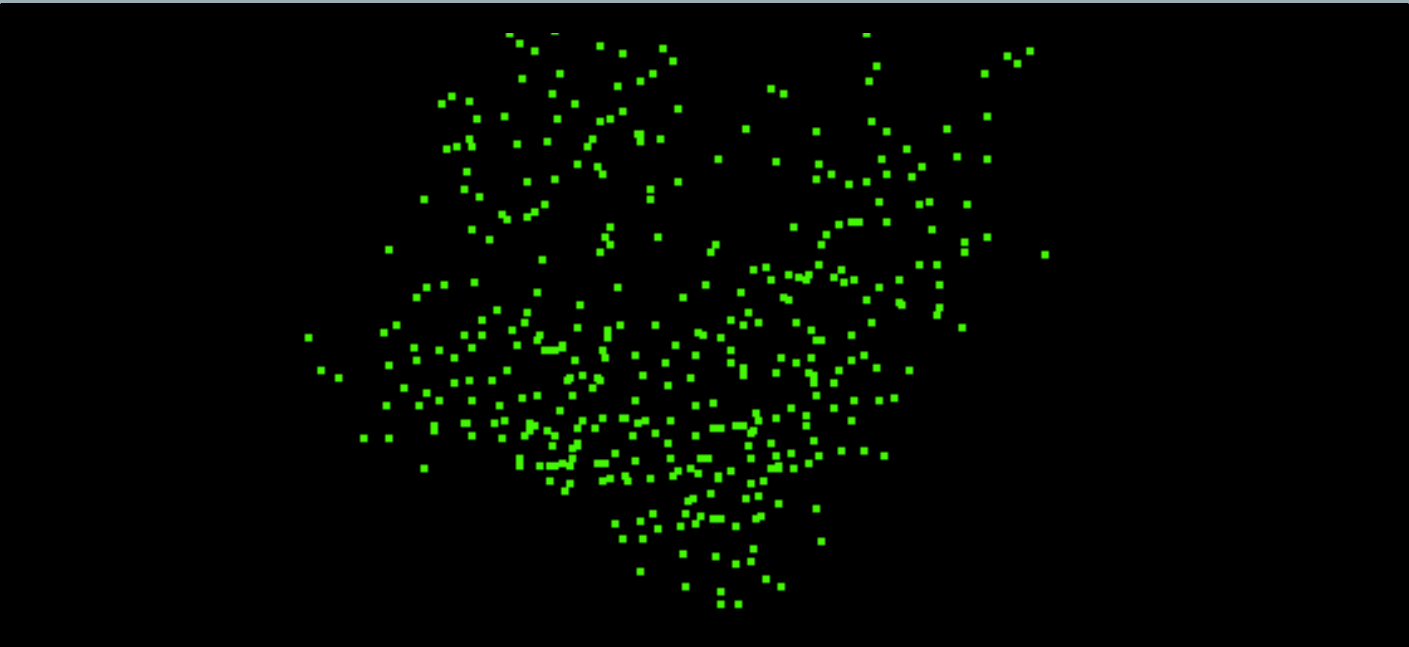} \\
\end{center}
\vspace{-6pt}
{\small
\hspace{40pt} (a) Quadric representation. \hspace{62pt}
(b) Plane representation. \hspace{66pt}
(c) Distribution representation.}
\captionof{figure}{DEPICTION OF VARIOUS WAYS TO REPRESENT 3D SCENES.}
\label{qpd}
\vspace{20pt}
}]

\section{More illustration of quadric Representations}
In this section, we provide additional illustrations of the three different representations of 3D scenes we use in our work, as enumerated in Fig.~\ref{qpd}.

\vspace{4pt}
\noindent \textbf{Quadric Representation.}
Referring to Sec.~3A, the coefficients of quadric functions $\mathbf{c}$ consist of two parts, the quadratic parts and linear parts. In practice, if any one of the quadric parts is not zero, the patches are quadric planes.

\vspace{4pt}
\noindent \textbf{Plane representation.} In practice, we find that fitting large ``patches'' such as walls and roads to normal quadric planes is computationally expensive and yields suboptimal performance in downstream tasks. In fact, a plane is a special case of a quadric where the quadratic coefficient $\boldsymbol{c}_q$ is set to zeros. To reduce the computation cost, we propose to use planes to represent large ``patches''. We first classify patch types according to a $3\times3$ covariance matrix, in which we perform eigenvalue decomposition for each patch. We consider patches with an eigenvalue much smaller than the other two eigenvalues as planes. Then, we assign the minimum eigenvector to the linear coefficient $\boldsymbol{c}_p$, and the constant term is assigned by easily calculating $\boldsymbol{c}_p$ and the center location $\boldsymbol{\mu}$ of the patch.

\vspace{4pt}
\noindent \textbf{Distribution representation.}
As illustrated in Sec.~3A, if the $mse$ of a patch is higher than a threshold, we represent such a patch as neither a quadric nor a plane. Instead, we consider it a degenerate plane, and all the coefficients of its quadric function are set to zero. In this case, we represent such a plane as a normal distribution.

For convenience, the aforementioned three classifications for patches will be called quadrics, planes, and normal distributions in the following illustrations. Each of these three different representations has a unique residual, namely point-to-quadratic, point-to-plane, and point-to-distribution distance, respectively. In addition, unless otherwise stated, our experimental results for odometry are scan-scan matching results. We use odometry experiments rather than mapping experiments for our ablation experiments because they are more sensitive to parameter adjustments, making it easier to observe the effect of each component of the system.

\section{Ablation Study on Odometry}

\begin{table*}[t]
    \centering
    \tabcolsep 8.8pt
    \caption{ODOMETRY RESULTS ON KITTI 00-10 SEQUENCES}  
    \vspace{-4pt}
    \begin{tabular}{p{2.5cm}<{\centering} | p{0.7cm}<{\centering} | p{0.5cm}<{\centering} p{0.5cm}<{\centering} p{0.5cm}<{\centering} p{0.5cm}<{\centering} p{0.5cm}<{\centering} p{0.5cm}<{\centering} p{0.5cm}<{\centering} p{0.5cm}<{\centering} p{0.5cm}<{\centering} p{0.5cm}<{\centering} p{0.5cm}<{\centering} | p{0.5cm}<{\centering}}
    \toprule 
    Method & Metric & 00U & 01H & 02C & 03C & 04C & 05C & 06U & 07U & 08U & 09C & 10C & Avg.\\
    \midrule
    D & \multirow{5.5}{*}{ATE} & 4.24 & 7.00 & 4.28 & 3.76 & 1.74 & 2.41 & 1.67 & 3.87 & 4.16 & 3.30 & 4.43 & 3.71 \\ 
    P + D &  & 2.73 & 10.13 & 3.91 & 2.07 & 2.18 & 2.91 & 1.68 & 2.56 & 2.97 & 3.15 & 5.00 & 3.57\\
    Q + D & & 3.08 & 10.65 & 2.27 & 1.76 & 2.04 & 2.31 & 1.43 & 2.78 & 3.16 & 2.30 & 3.75 & 3.23 \\
    Q + P + D (default)  & & 2.64 & \textbf{5.05} & 3.31 & \textbf{1.34} & \textbf{1.66} & \textbf{1.97} & \textbf{1.30} & 2.56 & 2.57 & 3.32 & \textbf{2.21} & \textbf{2.54}\\
    \cmidrule{1-1} 
    \cmidrule{3-14}
    Default + PC & & \textbf{1.74} & 7.92 & \textbf{2.42} & 2.67 & 2.12 & 2.23 & 1.37 & \textbf{1.93} & \textbf{1.77} & \textbf{2.21} & 3.10 & 2.68 \\
    \midrule
    \midrule
    D & \multirow{5.5}{*}{ARE} & 1.94 & 1.27 & 2.75 & 2.58 & 1.29 & 1.36 & 1.78 & 2.65 & 2.08 & 1.81 & 2.49 & 2.00 \\ 
    P + D & & 1.30 & 1.49 & 2.14 & 2.24 & 1.55 & 1.81 & 1.38 & 2.10 & 1.97 & 1.93 & 2.54 & 1.86 \\
    Q + D & & 1.44 & 1.37 & 2.34 & 2.10 & 1.54 & 1.53 & 2.28 & 1.48 & 1.58 & 1.60 & 2.42 & 1.79 \\
    Q + P + D (default) &  & 1.30 & 1.03 & 1.39 & \textbf{1.27} & 1.22 & 1.01 & \textbf{0.77} & 2.04 & 1.19 & 1.40 & \textbf{1.34} & 1.27\\
    \cmidrule{1-1} 
    \cmidrule{3-14}
    Default + PC &  & \textbf{0.76} & \textbf{0.87} & \textbf{0.93} & 1.56 & \textbf{1.00} & \textbf{1.00} & 0.77 & \textbf{1.02} & \textbf{0.94} & \textbf{1.07} & 1.79 & \textbf{1.06} \\
    \bottomrule
    \end{tabular}
    \begin{tablenotes}
        \centering
        \footnotesize
        \item ATE (\%), ARE (deg/100m), U: urban, H: highway, C: countryside. FULL: quadrics, planes and distributions. 
        P: planes. D: distribution. Q: quadrics. Dense: point-point residuals.
      \end{tablenotes}    
    \label{kitti_odo_abla}
\end{table*}

\subsection{Impact of Different Representations}
As explained in the above section, our method uses three different representations of 3D scenes: quadrics, planes, and normal distributions. Note that planes and normal distributions can be considered as two special cases of our proposed quadric representations. In this sub-section, we study how the different representations affect performance, as shown in Tab.~\ref{kitti_odo_abla}. Specifically, we investigate five cases:
\begin{itemize}
    \item Normal distributions (denoted as \textbf{D}). 
    We process all patches as a normal distribution. That is, given the points in a patch, we do not fit them into either patches or planes, but rather, we directly preserve the mean and the variance of these points.
    
    \item Planes and normal distributions (denoted as \textbf{P+D}). In this case, we first select the patches with $mse$ lower than the threshold and then classify them into ``plane'' and ``quadric'' patches. After that, we use normal distributions to represent the ``quadric'' patches rather than fitting them to quadric functions. We still represent patches with $mse$ higher than the threshold as normal distributions.
    
    \item Quadrics and normal distributions (denoted as \textbf{Q+D}). Similarly, after classifying the patches, we use normal distributions to represent       ``plane'' patches rather than linear functions, \textit{i.e.,} quadric functions the quadric coefficients equal to zeros. 
    
    \item  Quadrics, planes, and normal distributions (denoted as \textbf{Q+P+D} (default)). This is our default setting. For the patches with $mse$ lower than the threshold, we classify some as ``quadric'' patches and others as ``plane'' patches, using the process described above. Then, we fit them to quadric functions and linear functions, respectively. As for the patches with $mse$ higher than the threshold, we represent them as normal distributions.
    
    \item Besides exploring the combinations of quadrics, planes, and normal distributions, we also introduce point clouds into our method (denoted as \textbf{Default+PC}). Specifically, we preserve all points belonging to each patch along with the quadric representation. After quadric association, we additionally plug in the point-point distance loss, which is calculated as the distance between the points in the source patches and the corresponding nearest neighbors in the target patches.
\end{itemize}

\begin{figure}[t]
\centering
    \subfigure[Representing a quadric with a normal distribution.]{\includegraphics[height=1.4in]{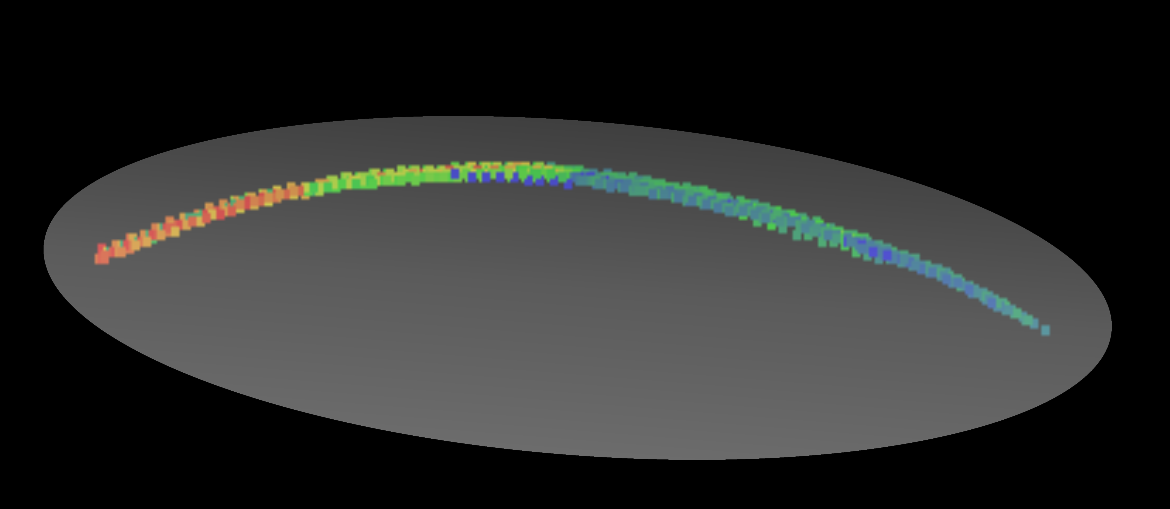}}
    \subfigure[Representing a plane with a normal distribution.]{\includegraphics[height=1.4in]{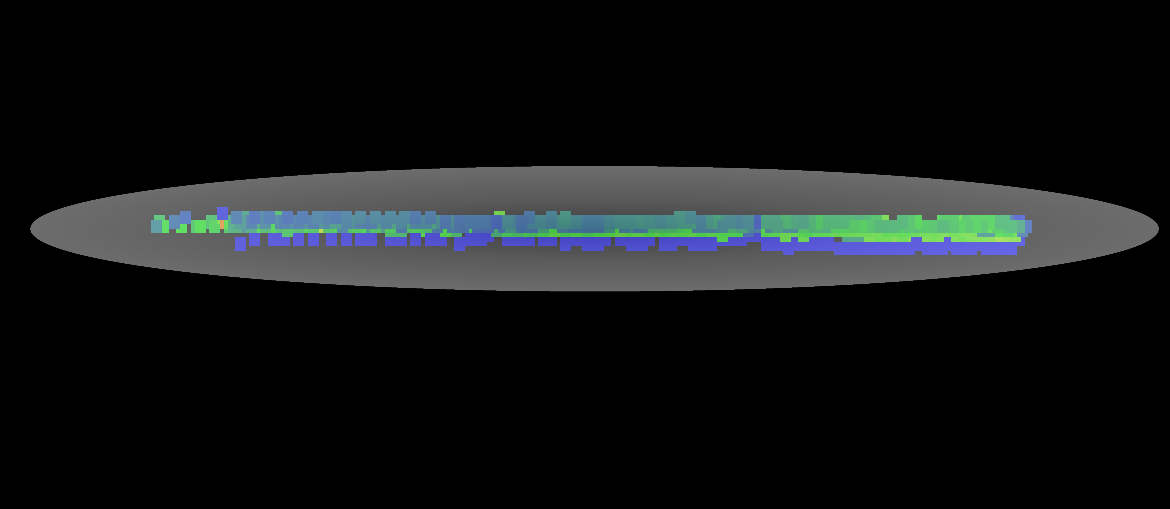}}
    \caption{ILLUSTRATION OF NORMAL DISTRIBUTIONS.}
    \label{distributions}
\end{figure}

\noindent\textbf{Normal distributions}. Using only normal distributions (\textbf{D}) for LiDAR odometry yields the worst performance of all representations, as shown in Tab.~\ref{kitti_odo_abla}. Indeed, the purely normal distribution representations of the scenes are prone to losing fine-grained information which is critical for scan-to-scan matching. As a result, the ATE and the ARE are higher than our default setting by 1.17 and 0.73, respectively. 

\noindent\textbf{Planes and normal distributions}. As we can see in Tab.~\ref{kitti_odo_abla}, using \textbf{P + D} to represent the 3D scenes achieves 3.57 ATE and 1.86 ARE, respectively. We observe that the accuracy drops by a large margin relative to \textbf{Q+D} (10.53\% for ATE, 3.91 deg/100m for ARE), and we hypothesize that normal distributions are a more compact alternative to planes compared to quadrics, as shown in Fig.~\ref{distributions}. Regarding the planes (Fig.~\ref{distributions}.(b)), an ellipse (normal distribution in 2D) can represent a plane well as long as the semiminor axis of the ellipse is short enough. Yet for a curved surface  (Fig.~\ref{distributions}.(a)), especially when the patch has a large surface area and a small curvature, it is hard for a normal-distribution ellipse to fit a curved surface (in 2D) well.

\noindent\textbf{Quadrics and normal distributions}.
As we can see in Tab.~\ref{kitti_odo_abla}, \textbf{Q + D} achieves 3.23 ATE and 1.79 ARE respectively. Although the performance of \textbf{Q + D} is better than \textbf{P + D}, we observe that the main performance gains are present in countryside scenarios. Indeed, different from the urban scenarios which are full of human-made ``Manhattan''-shaped objects, countryside scenarios have objects with more diverse shapes, thus yielding notable improvements when our method is applied to them.

\begin{table}[t]
    \centering
    \tabcolsep 5pt
    \caption{LATENCY (ms) COMPARISON BETWEEN FULL AND ADD RESULTS ON KITTI SEQUENCE 00}
    \begin{tabular}{c|c c c|c}
    \toprule
          Method & Feature Extraction & Association & Optimization & Total \\
          \midrule
         Default & 60.32 & 1.63 & 39.86 & 101.81 \\
         Default+PC & 60.32 & 31.27 & 88.05 & 179.64 \\
         \bottomrule
    \end{tabular}
    \label{time-consuming2}
\end{table}

\begin{figure*}[!htbp]
    \centering
    \subfigure[Trajectories.]{\includegraphics[width=0.32\linewidth]{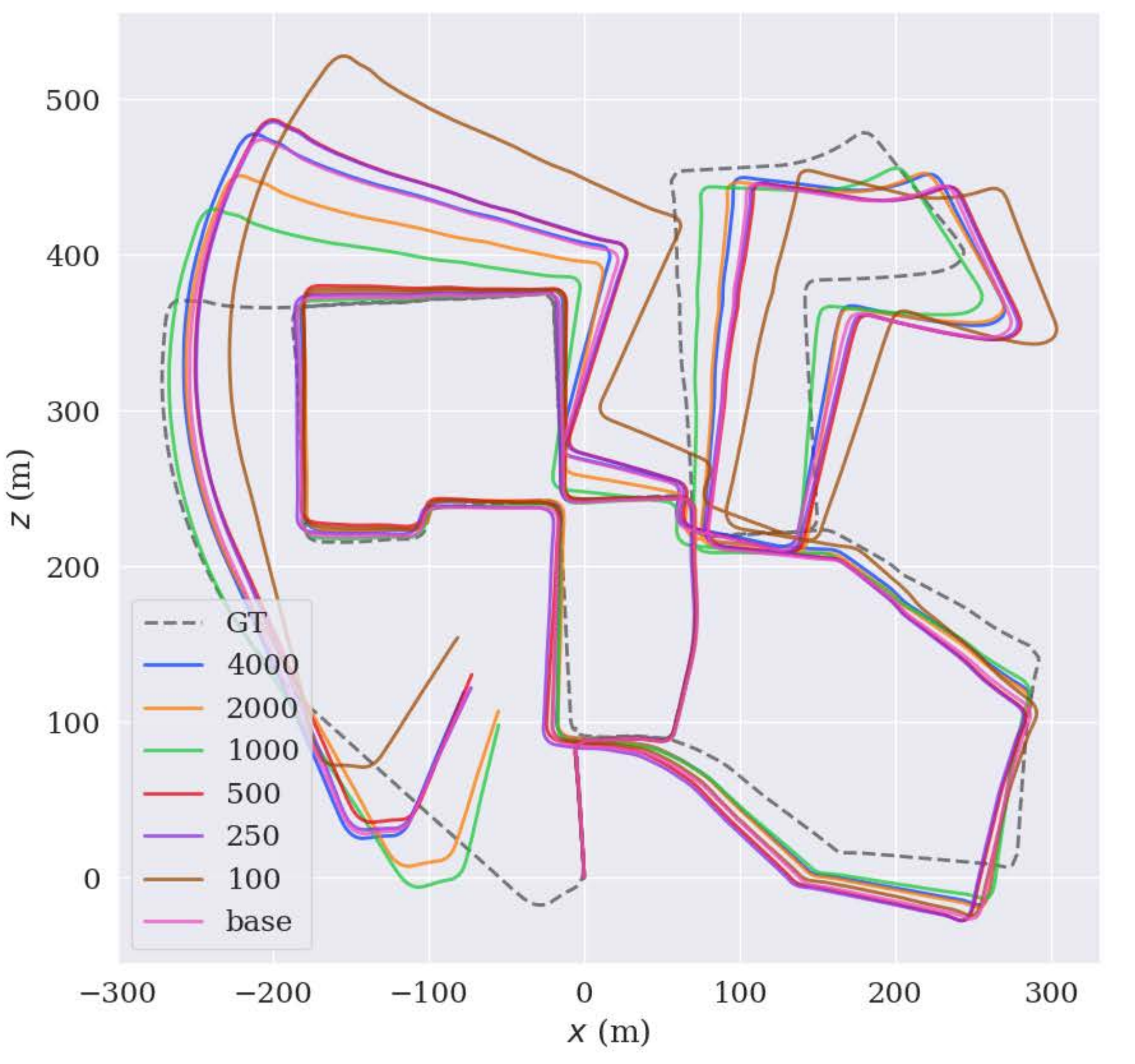}}
    \subfigure[ATE vs. Latency.]{
    \includegraphics[width=0.32\linewidth]{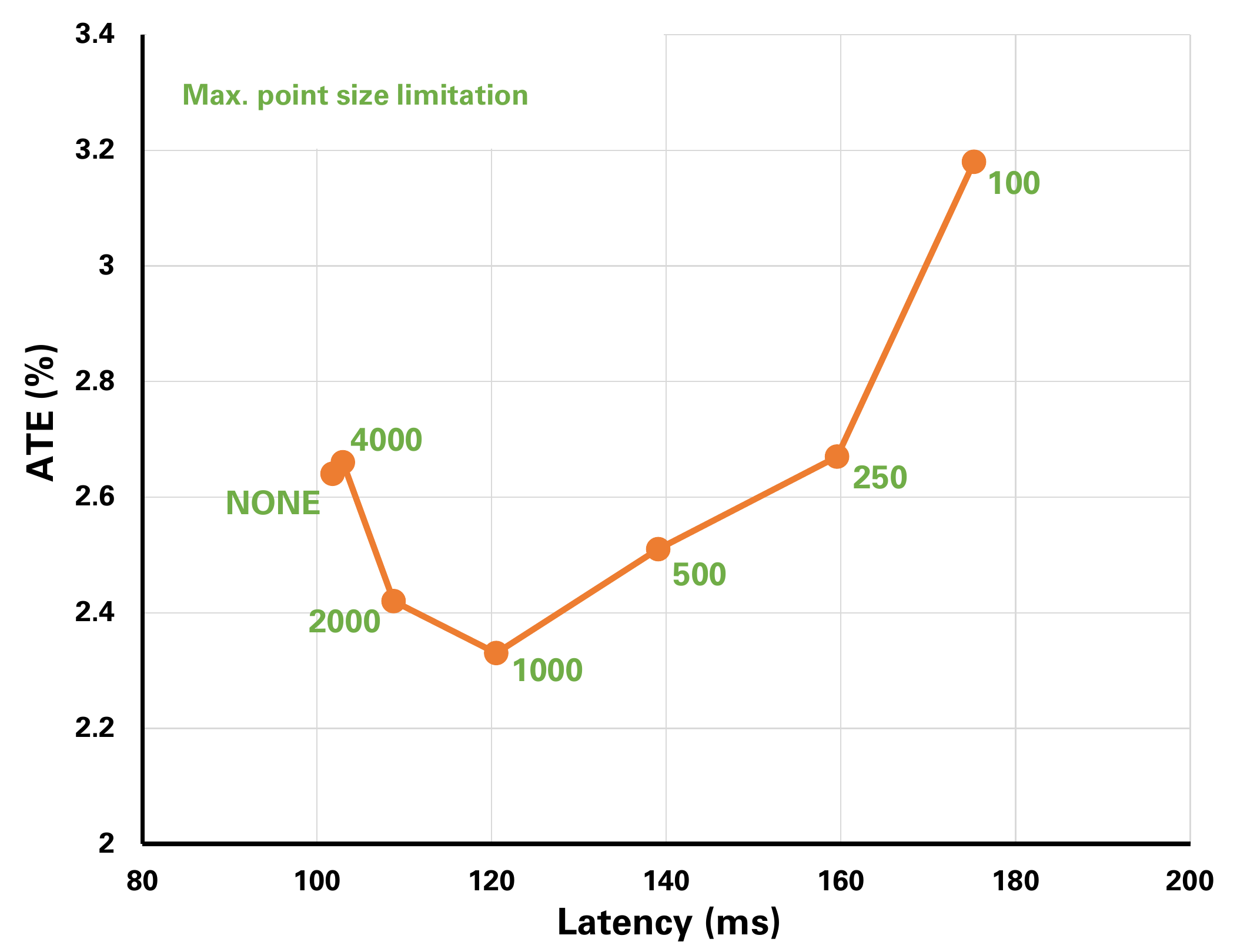}}
    \subfigure[ARE vs. Latency.]{
    \includegraphics[width=0.32\linewidth]{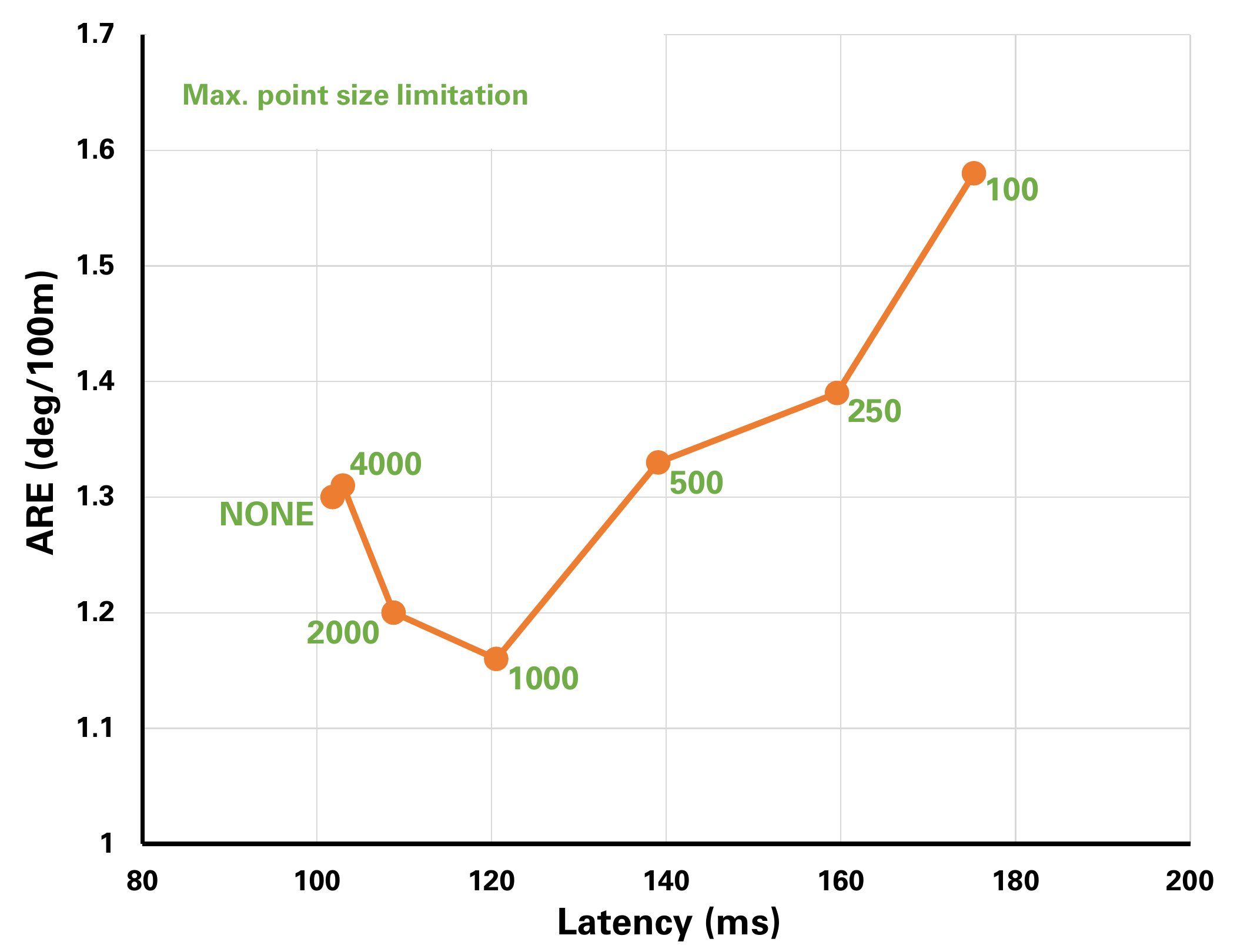}}
    \caption{Ablation experiment with different maximum point sizes in a patch.}
    \label{patch-size}
\end{figure*}

\begin{figure*}[t]
    \centering
    \includegraphics[width=0.32\linewidth, height=4cm]{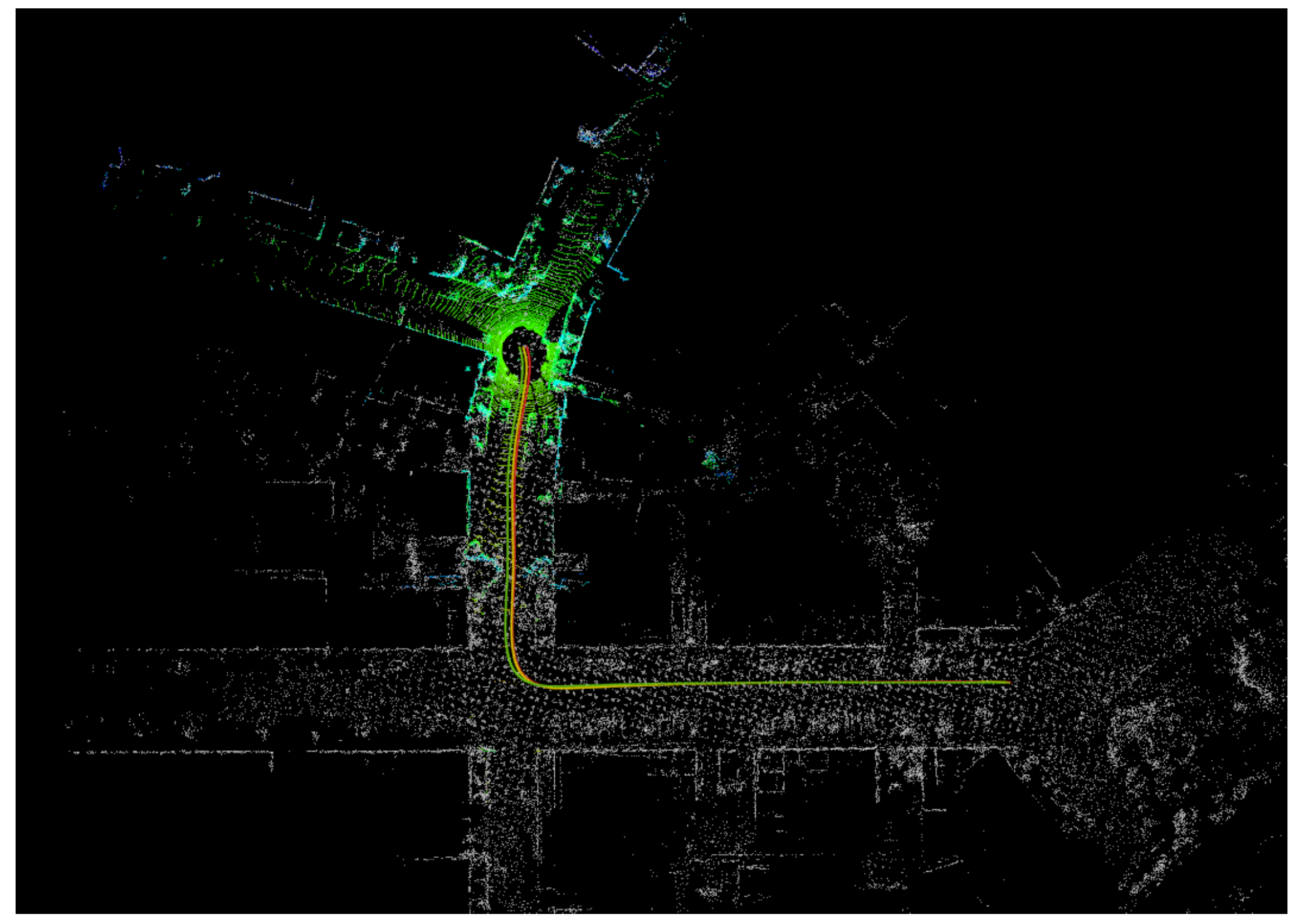}
    \includegraphics[width=0.32\linewidth, height=4cm]{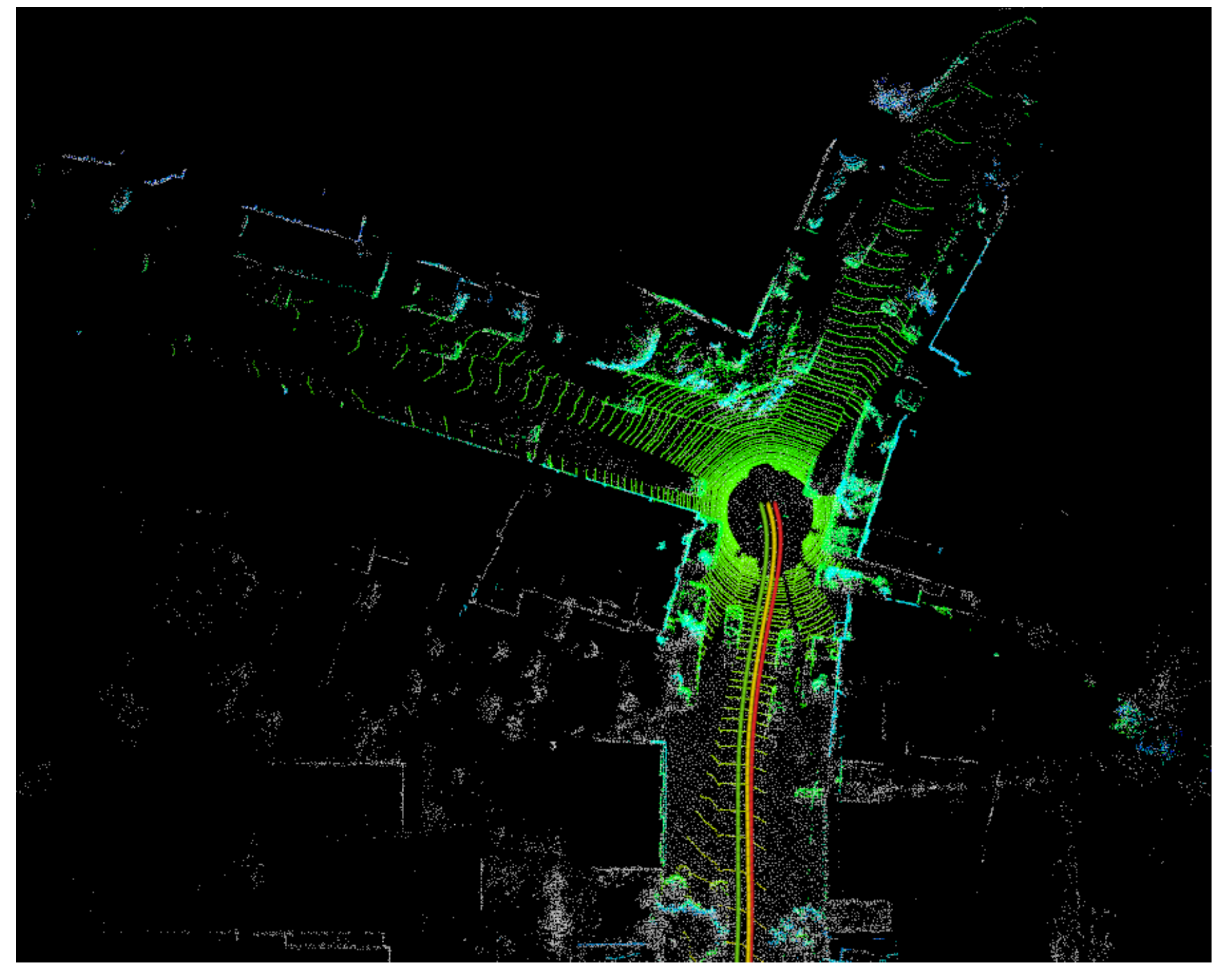}
    \includegraphics[width=0.32\linewidth, height=4cm]{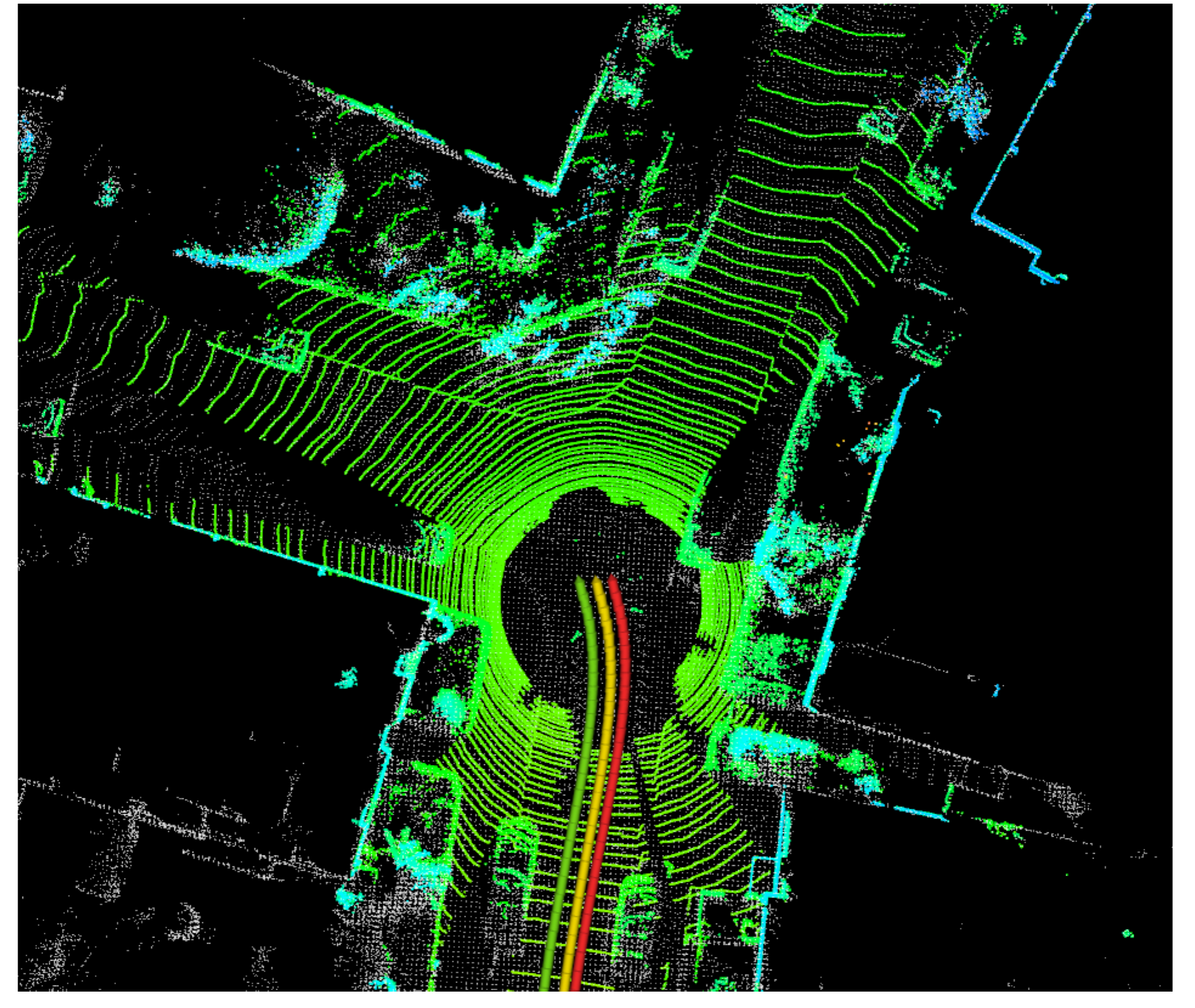}
    \caption{Visualization. White point cloud is map; Colored point cloud is real-time scan; Green trajectory is ground truth; Yellow trajectory is mapping result; Red trajectory is odometry result.}
    \label{visualization}
\end{figure*}

\noindent\textbf{Quadrics, planes, normal distributions, and point-clouds}.
It can be seen that after plugging in ``dense'' point-point residuals, both the ATE and ARE of the proposed method are lower when applied to urban scenes, while they are comparable or even higher when applied to countryside scenes. We hypothesize that the point-point correlation does not provide a completely correct gradient direction in countryside scenes due to the noisier environments with fewer structural features. This hypothesis also aligns with the results of the odometry experiment of ICP-po2po~\cite{besl1992method}. Moreover, the addition of dense points inevitably gives rise to much higher latency, as shown in Tab.~\ref{time-consuming2}. It can be observed that the main additional overhead comes from point-point association and optimization. Particularly, point-point association heavily relies on KDtree-based NNS, such that the computation cost is drastically increased. Meanwhile, due to the increasing residuals, the optimization of solving for optimal poses is much harder, resulting in further increased latency.

\subsection{Impact of Point Size} 
In this section, we investigate how different point size limitations affect performance. We define the point size of a patch as the number of points in it. We experiment with point size limitations of 100, 250, 500, 1000, 2000, and 4000 maximum points in a patch, and evaluate the corresponding performance on the KITTI 00 sequence.

Intuitively, the local fitting accuracy is higher when the point size is smaller, while the computation cost is also larger. We study the trade-off between the latency and the performance as the point size changes, as shown in Fig.~\ref{patch-size}. We observe that limiting the maximum point size to 4000 points leads to a slight drop in terms of both ATE and ARE compared to that when there is no maximum point size limitation. Many large patches in urban scenes, such as large walls and bushes, typically include around 5000 points. Thus, limiting the point size to 4000 may thus result in incomplete planes and undermine scan-to-scan matching accuracy.

Furthermore, as maximum point size decreases, performance gradually improves. The accuracy reaches its peak with a point size of 1000. However, a further decrease in the maximum point size from 1000 causes a gradual decline in performance.
This phenomenon can be attributed primarily to the patch segmentation process.
In this process, we segment a patch based on the geometric consistency between a point and its neighboring points.
When the number of points in a patch is artificially restricted, less geometric information is preserved. Thus, the shape of a patch is more likely to be randomly segmented, leading to a higher number of mismatches in the patch association module. 
To further improve performance, modifications need to be made to both the patch segmentation method and the patch selection process.

\noindent\textit{Visualization.} In Fig.~\ref{visualization}, we present trajectories from odometry and mapping experiments, along with the ground truth.

\vspace{12pt}

\bibliographystyle{plainnat}
\bibliography{IEEEabrv,ref}{}
\end{document}